\documentclass[10pt,journal,compsoc]{IEEEtran}
\usepackage[hyphens]{url}
\usepackage{hyperref}
\hypersetup{colorlinks=true,breaklinks=true}
\usepackage{cite}
\usepackage[pdftex]{graphicx}
\usepackage{amsmath}
\usepackage{amssymb}
\usepackage{mathtools}
\usepackage{algorithmic}
\usepackage[ruled, linesnumbered]{algorithm2e}
\usepackage{array}
\usepackage[caption=false]{subfig}
\usepackage{caption}
\usepackage{fixltx2e}
\usepackage{stfloats}
\usepackage{url}
\usepackage{multirow}
\usepackage{xcolor}
\usepackage[normalem]{ulem}
\usepackage{enumitem}
\usepackage{makecell}
\usepackage[normalem]{ulem}
\usepackage{xspace}
\usepackage{tabularx}
\usepackage{array, makecell} %
\usepackage{tablefootnote}
\usepackage{tabu}
\usepackage{fancyhdr}
\let\emptyset\varnothing

\DeclareMathOperator*{\argmin}{argmin}

\DeclareMathOperator{\Argmin}{argmin}
\DeclareMathOperator{\Argmax}{argmax}
\DeclareMathOperator{\Max}{max}
\DeclareMathOperator{\Min}{min}
\DeclareMathOperator{\Sup}{sup}
\usepackage{xspace}
\makeatletter
\DeclareRobustCommand\onedot{\futurelet\@let@token\@onedot}
\def\@onedot{\ifx\@let@token.\else.\null\fi\xspace}

\def\eg{\emph{e.g}\onedot} \def\Eg{\emph{E.g}\onedot}
\def\ie{\emph{i.e}\onedot}

\def\wrt{w.r.t\onedot} 
\def\etal{\emph{et al}\onedot}
\makeatother

\begin{document}

\title{Privacy-Preserving Deep Action Recognition: \\ An Adversarial Learning Framework \\and A New Dataset}
\author{Zhenyu~Wu*,
        Haotao~Wang*,
        Zhaowen~Wang,
        Hailin~Jin, 
        and~Zhangyang~Wang% <-this % stops a space
\IEEEcompsocitemizethanks{\IEEEcompsocthanksitem Zhenyu Wu is with the Department
of Computer Science and Engineering, Texas A\&M University, College Station,
TX, 77840. E-mail: wuzhenyu\_sjtu@tamu.edu 
\IEEEcompsocthanksitem Haotao Wang and Zhangyang Wang are with the Department
of Electrical and Computer Engineering, The University of Texas at Austin,
TX, 78712. E-mail: \{htwang,\,atlaswang\}@utexas.edu 
\IEEEcompsocthanksitem Zhaowen Wang and Hailin Jin are with Adobe Research, San Jose, CA, 95110. E-mail: \{zhawang, \,hljin\}@adobe.com
\IEEEcompsocthanksitem The first two authors Zhenyu Wu and Haotao Wang contributed equally to this work. 
\IEEEcompsocthanksitem  Correspondence to Zhangyang Wang (atlaswang@utexas.edu).
}% <-this % stops an unwanted space
}

\IEEEtitleabstractindextext{%
\begin{abstract}
We investigate privacy-preserving, video-based action recognition in deep learning, a problem with growing importance in smart camera applications. A novel adversarial training framework is formulated to learn an anonymization transform for input videos such that the trade-off between target utility task performance and the associated privacy budgets is explicitly optimized on the anonymized videos. Notably, the privacy budget, often defined and measured in task-driven contexts, cannot be reliably indicated using any single model performance because strong protection of privacy should sustain against any malicious model that tries to steal private information. To tackle this problem, we propose two new optimization strategies of model restarting and model ensemble to achieve stronger universal privacy protection against any attacker models. Extensive experiments have been carried out and analyzed.

\quad \quad
On the other hand, given few public datasets available with both utility and privacy labels, the data-driven (supervised) learning cannot exert its full power on this task. We first discuss an innovative heuristic of cross-dataset training and evaluation, enabling the use of multiple single-task datasets (one with target task labels and the other with privacy labels) in our problem. To further address this dataset challenge, we have constructed a new dataset, termed PA-HMDB51, with both target task labels (action) and selected privacy attributes (skin color, face, gender, nudity, and relationship) annotated on a per-frame basis. This first-of-its-kind video dataset and evaluation protocol can greatly facilitate visual privacy research and open up other opportunities. Our codes, models, and the PA-HMDB51 dataset are available at: {\emph{\url{https://github.com/VITA-Group/PA-HMDB51}}}.
\end{abstract}

\begin{IEEEkeywords}
Visual privacy, action recognition, privacy-preserving learning, adversarial learning.
\end{IEEEkeywords}}

\maketitle

\IEEEraisesectionheading{\section{Introduction}\label{sec:introduction}}
\IEEEPARstart{S}{mart} surveillance or smart home cameras, such as Amazon Echo and Nest Cam, are now found in millions of locations to link users to their homes or offices remotely. They provide the users with a monitoring service by notifying environment changes, a lifelogging service, and intelligent assistance. However, the benefits come at the heavy price of privacy intrusion from time to time.  Due to the computationally demanding nature of visual recognition tasks, only some of the tasks can run on the resource-limited local devices, which makes transmitting (part of) data to the cloud necessary. Growing concerns have been raised~\cite{smart-home-abuse2019vice, smart-security-camera2016forbes, doorbell-camera-surveillance2019washpost,media4, ban-video-surveillance2019forbes} towards personal data uploaded to the cloud, which could be potentially misused or stolen by malicious third-parties.
Many laws and regulations in the United States and the European Union~\cite{doj-552a-1974,weiss2016us,homeland-security2004-pnra,leenes2017data} also bring up guidelines for handling \emph{Personally Identifiable Information}.
This new privacy risk is fundamentally different from traditional concerns over unsecured transmission channels (\eg, malicious third-party eavesdropping), and therefore requires new solutions to address it. 

We ask if it is possible to alleviate privacy concerns without compromising user convenience. At first glance, the question itself is posed as a dilemma: we would like a camera system to recognize important events and assist daily human life by understanding its videos while preventing it from obtaining sensitive visual information (such as faces, gender, skin color, etc.) that can intrude individual privacy.
Thus, it becomes a new and appealing problem to find an appropriate transform to obfuscate the captured raw visual data at the local end, so that the transformed data will only enable specific target utility tasks while obstructing undesired privacy-related budget tasks.
Recently, some new video acquisition approaches~\cite{ryoo2017privacy,Butler2015ThePT,Dai2015TowardsPR} were proposed to intentionally capture or process videos in extremely low-resolution to create privacy-preserving ``anonymized'' videos and showed promising empirical results.

This paper takes {one of the first steps} towards addressing this {challenge} of privacy-preserving, video-based action recognition, via the following contributions: 
\begin{itemize}[leftmargin=*]
    \vspace{+0.25em}
    \item \textbf{A General Adversarial Training and Evaluation Framework.} 
    We address the privacy-preserving action recognition problem with a novel adversarial training framework. The framework explicitly optimizes the trade-off between target utility task performance and the associated privacy budgets by learning to anonymize the original videos.
    To reduce the training instability (as discussed in Section~\ref{sec:problem_def}), {we design and compare three different optimization strategies}. We empirically find one strategy generally outperforms the others under our framework and give intuitive explanations.
    
    \vspace{+0.25em}
    \item \textbf{Practical Approximations of ``Universal'' Privacy Protection.} The privacy budget in our framework cannot be defined \wrt~one model that predicts privacy attributes. Instead, the ideal protection of privacy must be universal and model-agnostic, \ie, preventing every possible attacker model from predicting private information. To resolve this so-called \textbf{``$\forall$ challenge''}, we propose two  effective strategies, \ie, \emph{restarting} and \emph{ensembling}, to enhance the generalization capability of the learned anonymization to defend against unseen models.
    We leave it as our future work to find better methods for this challenge.

    \vspace{+0.25em}
    \item \textbf{A New Dataset with Action and Privacy Annotations.} 
    When it comes to evaluating privacy protection on complicated privacy attributes, there is no off-the-shelf video dataset with both action (utility) and privacy attributes annotated, either for training or testing. Such a dataset challenge is circumvented in our previous work~\cite{wu2018towards} by using the VISPR~\cite{orekondy2017towards} dataset as an auxiliary dataset to provide privacy annotations for cross-dataset evaluation (details in Section 3.6). 
    However, this protocol inevitably suffers from the domain gap between the two datasets: while the utility was evaluated on one dataset, the privacy was measured on a different dataset. {The incoherence in utility and privacy evaluation datasets makes the obtained utility-privacy trade-off less convincing.} To reduce this gap, in this paper, we construct the very first testing benchmark dataset, dubbed \textbf{P}rivacy-\textbf{A}nnotated \textbf{HMDB51} (PA-HMDB51), to evaluate privacy protection and action recognition on the same videos simultaneously. 
    The new dataset consists of 515 videos originally from HMDB51. For each video, privacy labels (five attributes: skin color, face, gender, nudity, and relationship) are annotated on a per-frame basis. We benchmark our proposed framework on the new dataset and justify its effectiveness.
\end{itemize}
The paper is built upon our prior work~\cite{wu2018towards} with multiple improvements: (1) a detailed discussion and comparison on three optimization strategies for the proposed framework; (2) a more extensive experimental and analysis section; and (3) most importantly, the construction of the new PA-HMDB51 dataset, together with the associated benchmark results.
\section{Related Work}
\subsection{Privacy Protection in Computer Vision}
With pervasive cameras for surveillance or smart home devices, privacy-preserving action recognition has drawn increasing interests from both industry and academia. 

\vspace{+0.25em}
\noindent \textbf{Transmitting Feature Descriptors.} 
A seemingly reasonable and computationally cheaper option is to extract feature descriptors from raw images and transmit those features only. Unfortunately, previous studies~\cite{pittaluga2019revealing,dosovitskiy2016inverting,kato2014image,weinzaepfel2011reconstructing,Mahendran_2016} revealed that considerable details of original images could still be recovered from standard HOG, SIFT, LBP, 3D point clouds, Bag-of-Visual-Words or neural network activations (even if they look visually distinctive from natural images).

\vspace{+0.25em}
\noindent \textbf{Homomorphic Cryptographic Solutions.} 
Most classical cryptographic solutions secure communication against unauthorized access from attackers. However, they are not immediately applicable to preventing authorized agents (such as the back-end analytics) from the unauthorized abuse of information, causing privacy breach concerns. A few encryption-based solutions, such as Homomorphic Encryption (HE)~\cite{gentry2009fully,xie2014crypto}, were developed to locally encrypt visual information. The server can only get access to the encrypted data and conduct a utility task on it. 
However, many encryption-based solutions will incur high computational costs at local platforms. 
It is also challenging to generalize the cryptosystems to more complicated classifiers. Chattopadhyay~\etal~\cite{chattopadhyay2007privacycam} combined the detection of regions of interest and the real encryption techniques to improve privacy while allowing general surveillance to continue. 

\vspace{+0.25em}
\noindent \textbf{Anonymization by Empirical Obfuscations.}
An alternative approach towards a privacy-preserving vision system is based on the concept of anonymized videos. 
Such videos are intentionally captured or processed by empirical obfuscations to be in special low-quality conditions, which only allow for recognizing some target events or activities while avoiding the unwanted leak of the identity information for the human subjects in the video.

Ryoo~\etal~\cite{ryoo2017privacy} showed that even at the extremely low resolutions, reliable action recognition could be achieved by learning appropriate downsampling transforms, with neither unrealistic activity-location assumptions nor extra specific hardware resources. The authors empirically verified that conventional face recognition easily failed on the generated low-resolution videos.
Butler~\etal~\cite{Butler2015ThePT} used image operations like blurring and superpixel clustering to get anonymized videos, while Dai~\etal~\cite{Dai2015TowardsPR} used extremely low resolution (\eg, $16 \times 12$) camera hardware to get anonymized videos.
Winkler~\etal~\cite{winkler2014trusteye} used cartoon-like effects with a customized version of mean shift filtering.
Wang~\etal~\cite{wang2019privacy} proposed a lens-free coded aperture (CA) camera system, producing visually unrecognizable and unrestorable image encodings. Pittaluga \& Koppal
\cite{pittaluga2015privacy,pittaluga2017pre} proposed to use privacy-preserving optics to filter sensitive information from the incident light-field before sensor measurements are made, by $k$-anonymity and defocus blur. Earlier work of Jia~\etal~\cite{jia2014using} explored privacy-preserving tracking and coarse pose estimation using a network of ceiling-mounted time-of-flight low-resolution sensors. Tao~\etal~\cite{tao2012privacy} adopted a network of ceiling-mounted binary passive infrared sensors.
However, both works~\cite{jia2014using,tao2012privacy} handled only a limited set of activities performed at specific constrained areas in the room.

The usage of low-quality anonymized videos by obfuscations was computationally cheap and compatible with sensor's bandwidth constraints. However, the proposed obfuscations were not learned towards protecting any visual privacy, thus having limited effects. In other words, privacy protection came as a ``side product'' of obfuscation, and was not a result of any optimization, making the privacy protection capability very limited. What is more, the privacy-preserving effects were not carefully analyzed and evaluated by human study or deep learning-based privacy recognition approaches. Lastly, none of the aforementioned empirical obfuscations extended their efforts to study deep learning-based action recognition, making their task performance less competitive. Similarly, the recent progress of low-resolution object recognition~\cite{Wang2016StudyingVL,cheng2017robust,xu2018fully} also put their privacy protection effects in jeopardy. 

\vspace{+0.25em}
\noindent \textbf{Learning-based Solutions.}
Very recently, a few learning-based approaches have been proposed to address privacy protection or fairness problems in vision-related tasks~\cite{wu2018towards,wu2019delving,m2019all,pittaluga2019learning,bertran2019adversarially,roy2019mitigating,zhang2018mitigating,ren2018learning,shetty2018adversarial,wang2019balanced}. Many of them exploited ideas from adversarial learning. They addressed this problem by learning data representations that simultaneously reduce the budget cost of privacy or fairness while maintaining the utility task performance. 

Wu~\etal~\cite{wu2019delving} proposed an adversarial training
framework dubbed Nuisance Disentangled Feature Transform (NDFT) to utilize the free meta-data (\ie, altitudes, weather conditions, and viewing angles) in conjunction with associated UAV images to learn domain-robust features for object detection in UAV images.
Pittaluga~\etal~\cite{pittaluga2019learning} preserved the utility by maintaining the variance of the encoding or favoring a second classifier for a different attribute in training. Bertran~\etal~\cite{bertran2019adversarially} motivated the adversarial learning framework as a distribution matching problem and defined the objective and the constraints in mutual information. Roy \& Boddeti~\cite{roy2019mitigating} measured the uncertainty in the privacy-related attributes by the entropy of the discriminator's prediction.
Oleszkiewicz~\etal~\cite{oleszkiewicz2018siamese} proposed an empirical data-driven privacy metric based on mutual information to quantify the privatization effects on biometric images.
Zhang~\etal~\cite{zhang2018mitigating} presented an adversarial debiasing framework to mitigate the biases concerning demographic groups. Ren~\etal~\cite{ren2018learning} learned a face anonymizer in video frames while maintaining the action detection performance. Shetty~\etal~\cite{shetty2018adversarial} presented an automatic object removal model that learns how to find and remove objects from general scene images via a generative adversarial network (GAN) framework.

\subsection{Privacy Protection in Social Media/Photo Sharing}
User privacy protection is also a topic of extensive interest in the social media field, especially for photo sharing. The most common means to protect user privacy in an uploaded photo is to add empirical obfuscations, such as blurring, mosaicing, or cropping out certain regions (usually faces)~\cite{li2017blur}. However, extensive research showed that such an empirical approach could be easily hacked~\cite{oh2016faceless,mcpherson2016defeating}.
A recent work~\cite{oh2017adversarial} described a game-theoretical system in which the photo owner and the recognition model strive for antagonistic goals of disabling recognition, and better obfuscation ways could be learned from their competition. However, their system was only designed to confuse one specific recognition model via finding its adversarial perturbations.
Fooling only one recognition model can cause obvious overfitting as merely changing to another recognition model will likely put the learning efforts in vain: such perturbations cannot even protect privacy from human eyes. The problem setting in~\cite{oh2017adversarial} thus differs from our target problem.
Another notable difference is that we usually hope to generate minimum perceptual quality loss to photos after applying any privacy-preserving transform to them in social photo sharing. There is no such restriction in our scenario. We can apply a much more flexible and aggressive transformation to the image.

The visual privacy issues faced by blind people were revealed in~\cite{gurari2019vizwiz} with the first dataset in this area.
Concrete privacy attributes were defined in~\cite{orekondy2017towards} with their correlation with image content. The authors categorized possible private information in images, and they ran a user study to understand privacy preferences. 
They then provided a sizable set of 22k images annotated with 68 privacy attributes, on which they trained privacy attributes predictors.
\section{Method}

\subsection{Problem Definition} \label{sec:problem_def}
\vspace{+0.25em}
\noindent\textbf{Objective.} 
Assume our training data $X$ (raw visual data captured by camera) are associated with a target utility task $\mathcal T$ and a privacy budget $\mathcal{B}$.
Since $\mathcal{T}$ is usually a supervised task, \eg, action recognition or visual tracking, a label set $Y_T$ is provided on $X$, and a standard cost function $L_T$ (\eg, cross-entropy) is defined to evaluate the task performance on $\mathcal{T}$. Usually, there is a state-of-the-art deep neural network $f_T$, which takes $X$ as input and predicts the target labels.
On the other hand, we need to define a budget cost function $J_B$ to evaluate its input data's privacy leakage: the smaller $J_B(\cdot)$ is, the less private information its input contains.

We seek an optimal anonymization function $f_A^*$ to transform the original $X$ to anonymized visual data $f_A^*(X)$, and an optimal target model $f_T^*$ such that:
\begin{itemize}[leftmargin=*]
\item $f_A^*$ has filtered out the private information in $X$, \ie, \\
\begin{equation*} %\label{eq:condition_b}
J_B(f_A^*(X)) \ll J_B(X);
\end{equation*}
\item the performance of $f_T$ is minimally affected when using the anonymized visual data $f_A^*(X)$ compared to when using the original data $X$, \ie, \\
\begin{equation*} %\label{eq:condition_b}
L_T(f_T^*(f_A^*(X)),Y_T)  \approx \Min_{f_T} L_T(f_T(X),Y_T).
\end{equation*}
\end{itemize}

To achieve these two goals, we mathematically formulate the problem as solving the following optimization problem:
\begin{align} 
\begin{split} \label{eq:loss-old}
f_A^*, f_T^* = \argmin_{f_A,f_T}[L_T(f_T(f_A(X)),Y_T) + \gamma J_B(f_A(X))].
\end{split}
\end{align}

\vspace{+0.25em}
\noindent \textbf{Definition of $J_B$ and $L_T$.} 
The definition of the privacy budget cost $J_B$ is not straightforward. Practically, it needs to be placed in concrete application contexts, often in a task-driven way. For example, in smart workplaces or smart homes with video surveillance, one might often want to avoid disclosure of the face or identity of persons. Therefore, to reduce $J_B$ could be interpreted as to suppress the success rate
of identity recognition or verification. Other privacy-related attributes, such as race, gender, or age, can be similarly defined too. 
We denote the privacy-related annotations (such as identity label) as $Y_B$, and rewrite $J_B(f_A(X))$ as $J_B(f_B(f_A(X)),Y_B)$, where $f_B$ denotes the privacy budget model which takes (anonymized or original) visual data as input and predicts the corresponding private information. Different from $L_T$, minimizing $J_B$ will encourage $f_B(f_A(X))$ to diverge from $Y_B$.
Without loss of generality, we assume both $f_T$ and $f_B$ to be classification models and output class labels. 
% Under this assumption, we could choose both $L_T$ and $L_B$ as the cross-entropy function, and $J_B$ as the negative cross-entropy function:
% \begin{equation*}
%     J_B  \triangleq -H(Y_B, f_B(f_A(X)))
% \end{equation*}
% where $H(\cdot,\cdot)$ is the cross-entropy function. 
Under this assumption, we choose both $L_T$ and $L_B$ as the cross-entropy function, and $J_B$ as the negative cross-entropy function:
\begin{align*}
    J_B  &\triangleq -H(Y_B, f_B(f_A(X))),
\end{align*}
where $H(\cdot,\cdot)$ is the cross-entropy function. 
% For convenience, we also define another variable $L_B$ as $-J_B$:
% \begin{align*}
%     L_B \triangleq -J_B = H(Y_B, f_B(f_A(X))).
% \end{align*}

\vspace{+0.25em}
\noindent \textbf{Two Challenges.} 
Such a supervised, task-driven definition of $J_B$ poses at least two challenges: 
(1)\ \emph{Dataset challenge:} 
The privacy budget-related annotations, denoted as $Y_B$, often have less availability than target utility task labels. Specifically, it is often challenging to have both $Y_T$ and $Y_B$ available on the same $X$; 
(2)\ \emph{$\boldsymbol \forall$ challenge:} 
Considering the nature of privacy protection, it is not sufficient to merely suppress the success rate of one $f_B$ model. Instead, we define a privacy prediction function family 
\begin{equation*}
    \mathcal{P}: f_A(X) \mapsto Y_B,
\end{equation*}
so that the ideal privacy protection by $f_A$ should be reflected as \textit{suppressing every possible model} $f_B$ from $\mathcal{P}$. That differs from the common supervised training goal, where only one model needs to be found to fulfill the target utility task successfully. 

We address the \emph{dataset challenge} by two ways: (1) cross dataset training and evaluation (Section~\ref{sec:cross-dataset}); and more importantly (2) building a new dataset annotated with both utility and privacy labels (Section~\ref{sec:dataset}). We defer their discussion to respective experimental paragraphs. 

Handling the \emph{$\boldsymbol \forall$ challenge} is more challenging.
Firstly, we re-write the general form in Eq.~(\ref{eq:loss-old}) with the task-driven definition of $J_B$ as follows:
\begin{align} 
\begin{split} \label{eq:loss-sup}
f_A^*, f_T^* = \Argmin_{(f_A,f_T)}&[L_T(f_T(f_A(X)),Y_T) + \\
\gamma & \Sup_{f_B\in\mathcal{P}}J_B(f_B(f_A(X)),Y_B)].
\end{split}
\end{align}
The \emph{$\boldsymbol \forall$ challenge} is the infeasibility to directly solve Eq.~(\ref{eq:loss-sup}), due to the infinite search space of $f_B$ in $\mathcal{P}$. Secondly, we propose to solve the following approximate problem by setting $f_B$ as a neural network with a fixed structure:
\begin{align} 
\begin{split} \label{eq:loss-max}
f_A^*, f_T^* = \Argmin_{(f_A,f_T)}&[L_T(f_T(f_A(X)),Y_T) + \\
\gamma & \Max_{f_B}J_B(f_B(f_A(X)),Y_B)].
\end{split}
\end{align}
Lastly, we propose ``model ensemble'' and ``model restarting'' (Section~\ref{sec:address-any}) to handle the \emph{$\boldsymbol \forall$ challenge} better and boost the experimental results further.

Considering the \emph{$\boldsymbol \forall$ challenge}, the evaluation protocol for privacy-preserving action recognition is more intricate than traditional action recognition task. We propose a two-step protocol (as described in Section~\ref{sec:two-step}) to evaluate $f_A^*$ and $f_T^*$ on the trade-off they have achieved between target task utility and privacy protection budget.

\vspace{+0.25em}
\noindent \textbf{Solving the Minimax.} 
Solving Eq.~(\ref{eq:loss-max}) is still challenging because the minimax problem is hard by its nature. {Traditional minimax optimization algorithms based on alternating gradient descent can only find minimax points for convex-concave problems, and they achieve sub-optimal solutions on deep neural networks since they are neither convex nor concave.}
Some very recent minimax algorithms, such as $K$-Beam~\cite{Hamm2018KBeamME}, have been shown to be promising in none convex-concave and deep neural network applications. However, these methods rely on heavy parameter tuning and are effective only in limited situations.
Besides, our optimization goal in Eq.~(\ref{eq:loss-max}) is even harder than common minimax objectives like those in GANs, which are often interpreted as a two-party competition game. In contrast, our Eq.~(\ref{eq:loss-max}) is more ``hybrid'' and can be interpreted as a more complicated three-party competition, where (adopting machine learning security terms) $f_A$ is an obfuscator, $f_T$ is a utilizer collaborating with the obfuscator, and $f_B$ is an attacker trying to breach the obfuscator. Therefore, we see no obvious best choice from the off-the-shelf minimax algorithms to achieve our objective.

We are thus motivated to try different state-of-the-art minimax optimization algorithms on our framework. 
We tested two state-of-the-art minimax optimization algorithms, namely GRL~\cite{ganin2014unsupervised} and $K$-Beam~\cite{Hamm2018KBeamME}, on our framework and proposed an innovative 
% method dubbed maximize entropy (ME)
entropy maximization method to solve Eq.~(\ref{eq:loss-max}). 
We empirically show our entropy maximization algorithm outperforms both state-of-the-art minimax optimization algorithms and discuss its advantages. 
In Section~\ref{sec:algos}, we present the comparison of three methods and hope it will benefit future research on similar problems.

\subsection{Basic Framework} \label{sec:framework}
\begin{figure}[!t]
\centering
\includegraphics[width=\columnwidth]{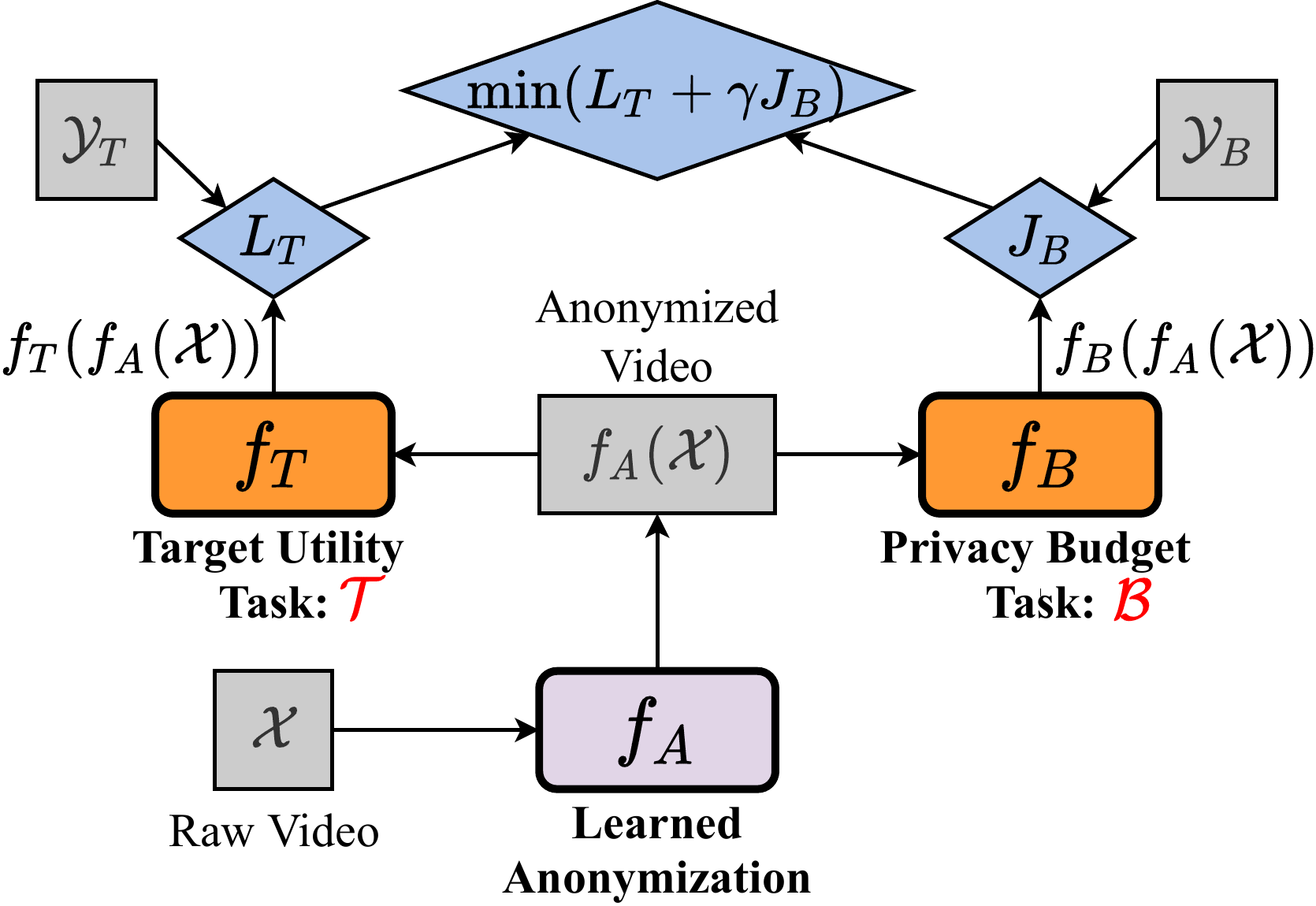}
\vspace{-1em}
\caption{Basic adversarial training framework for privacy-preserving action recognition.}
\label{fig:framework}
\end{figure}

\vspace{+0.25em}
\noindent\textbf{Pipeline.} 
Our framework is a privacy-preserving action recognition pipeline that uses video data as input. It is a prototype of the in-demand privacy protection in smart camera applications. Figure~\ref{fig:framework} depicts the basic framework implementing the proposed formulation in Eq.~(\ref{eq:loss-max}).
The framework consists of three parts: the anonymization model $f_A$, the target utility model $f_T$, and the privacy budget model $f_B$. $f_A$ takes raw video $X$ as input, filters out private information in $X$, and outputs the anonymized video $f_A(X)$. 
$f_T$ takes $f_A(X)$ as input and carries out the target utility task. $f_B$ also take $f_A(X)$ as input and try to predict the private information from $f_A(X)$.
All three models are implemented with deep neural networks, and their parameters are learnable during the training procedure. 
The entire pipeline is trained under the guidance of the hybrid loss of $L_T$ and $J_B$.
The training procedure has two goals. The first goal is to find an optimal anonymization model $f_A^*$ that can filter out the private information in the original video while keeping useful information for the target utility task. The second goal is to find a target model that can achieve good performance on the target utility task using anonymized videos $f_A^*(X)$. Similar frameworks have been used in feature disentanglement~\cite{xiang2018linear, desjardins2012disentangling, gonzalez2018image, reddy2016unbounded}.
After training, the learned anonymization model can be applied on a local device (\eg, smart camera), by designing an embedded chipset
responsible for the anonymization at the hardware-level~\cite{ren2018learning}. We can convert raw video to anonymized video locally and only transfer the anonymized video through the Internet to the backend (\eg, cloud) for target utility task analysis. The private information in the raw videos will be unavailable on the backend.

\vspace{+0.25em}
\noindent\textbf{Implementation.} 
Specifically, $f_A$ is implemented using the model in~\cite{Johnson2016PerceptualLF}, which can be taken as a 2D convolution-based frame-level filter. In other words, $f_A$ converts each frame in $X$ into a feature map of the same shape as the original frame.
We use state-of-the-art human action recognition model C3D\cite{tran2015learning} as $f_T$ and state-of-the-art image classification models, such as ResNet\cite{He_2016} and MobileNet\cite{howard2017mobilenets}, as $f_B$.
Since the action recognition model we use is C3D, we need to split the videos into clips with a fixed frame number. Each clip is a 4D tensor of shape $[T,W,H,C]$, where $T$ is the number of frames in each clip and $W$, $H$, $C$ are the width, height, and the number of color channels in each frame respectively. 
Unlike $f_T$, which takes a 4D tensor as an input data sample, $f_B$ takes a 3D tensor (\ie, a frame) as input. We average\footnote{ AVERAGING the logits temporally gave a better performance in privacy budget prediction of $J_B$, compared with MAXING the logits.} the logits over the temporal dimension of each video clip to calculate $J_B$ and predict the budget task label.

\subsection{Optimization Strategies} \label{sec:algos}
In the following algorithms, we denote $\theta_B$ as the parameters of $f_B$. Similarly, $f_A$ and $f_T$ are parameterized by $\theta_A$ and $\theta_T$ respectively. $\alpha_A, \alpha_B, \alpha_T$ are learning rates used to update $\theta_A, \theta_B, \theta_T$. $th_B$ and $th_T$ are accuracy thresholds for the target utility task and the privacy budget prediction. $max\_iter$ is the maximum number of iterations.
For simplicity concern, we abbreviate $L_T(f_T(f_A(X)), Y_T)$, $L_B(f_B(f_A(X)), Y_B)$, and $J_B(f_B(f_A(X)), Y_B)$ as $L_T(\theta_A, \theta_T)$, $L_B(\theta_A, \theta_B)$, and $J_B(\theta_A, \theta_B)$\footnote{Remember that $J_B$ is the negative cross-entropy by definition.} respectively. $Acc$ is a function to compute accuracy on the privacy budget and the target utility tasks, given training data ($X^t$, $Y^t_B$ and $Y^t_T$) or validation data ($X^v$, $Y^v_B$ and $Y^v_T$).

\subsubsection{Gradient reverse layer (GRL)} \label{sec:grl}
We consider Eq.~(\ref{eq:loss-max}) as a minimax problem ~\cite{du2013minimax}:

\begin{align*}
\theta_A^*,\theta_T^* &= \Argmin_{(\theta_A,\theta_T)} L(\theta_A, \theta_T, \theta_B^*), \\
\theta_B^* &= \Argmax_{\theta_B} L(\theta_A^*, \theta_T^*, \theta_B),
\end{align*}

where $L(\theta_A, \theta_T, \theta_B)=L_T(\theta_A, \theta_T)+\gamma J_B(\theta_A, \theta_B)=L_T(\theta_A, \theta_T)-\gamma L_B(\theta_A, \theta_B)$.

GRL\cite{ganin2014unsupervised} is a state-of-the-art algorithm to solve such a minimax problem. 
The underlying mathematical gist is to solve the problem by alternating minimization:
\begin{subequations} \label{eq:GRL-loss}
\begin{align} % & indicates where to align
\theta_A &\gets \theta_A - \alpha_A \nabla_{\theta_A} (L_T(\theta_A,\theta_T) - \gamma L_B(\theta_A, \theta_B)), \label{eq:GRL-D} \\
\theta_T &\gets \theta_T - \alpha_T \nabla_{\theta_T} L_T(\theta_A, \theta_T), \label{eq:GRL-T} \\
\theta_B &\gets \theta_B - \alpha_B \nabla_{\theta_B} L_B(\theta_A, \theta_B).  \label{eq:GRL-B}
\end{align}
\end{subequations}

We denote this method as {\textbf{GRL}} in the following parts and give the details in Algorithm~\ref{alg:GRL}.

\begin{algorithm}
\SetAlgoLined
Initialize $\theta_A$, $\theta_{T}$ and $\theta_{B}$\;
\For{$t_0 \gets 1$ \KwTo max\_iter}
{
    Update $\theta_A$ using Eq.~(\ref{eq:GRL-D}) \\
    \While{$Acc(f_T(f_A(X^v)),Y_T^v)$ $\leq th_T$ }
    {
        Update $\theta_T$ using Eq.~(\ref{eq:GRL-T})
    }
    \While{$Acc(f_B(f_A(X^t)),Y_B^t)$ $\leq th_B$}
    {
        Update $\theta_{B}$ using Eq.~(\ref{eq:GRL-B})
    }
}
\caption{GRL Algorithm}
 \label{alg:GRL}
\end{algorithm}

\subsubsection{Alternating optimization of two loss functions} \label{sec:kbeam}
The goal in Eq.~(\ref{eq:loss-max}) can also be formulated as alternatingly solving the following two optimization problems:
\begin{subequations} \label{eq:kbeam-loss}
\begin{align}
\theta_A^*, \theta_T^* &= \Argmin_{(\theta_A,\theta_T)} \  L_T(\theta_A,\theta_T), \label{eq:kbeam-min} \\
\theta_B^*, \theta_A^* &= \Argmin_{\theta_B} \Argmax_{\theta_A} \  L_B(\theta_A, \theta_B). \label{eq:kbeam-minimax}
\end{align}
\end{subequations}
Eq.~(\ref{eq:kbeam-min}) is a standard minimization problem which can be solved by end-to-end training $f_A$ and $f_T$. Eq.~(\ref{eq:kbeam-minimax}) is a minimax problem which we solve by state-of-the-art minimax optimization method ``$K$-Beam''~\cite{Hamm2018KBeamME}. 
$K$-Beam method keeps track of $K$ different sets of budget model parameters (denoted as $\{\theta_B^i\}_{i=1}^K$) during training time, and alternatingly updates $\theta_A$ and $\{\theta_B^i\}_{i=1}^K$. 

Inspired by $K$-Beam method, we present the following parameter update rules to alternatingly solve the two loss functions in Eq.~(\ref{eq:kbeam-loss}):
\begin{subequations} \label{eq:kbeam-update}
\begin{align}
&\theta_A,\theta_T \gets \theta_A, \theta_T - \alpha_T \nabla_{(\theta_T, \theta_A)} L_T(\theta_A,\theta_T), \label{eq:kbeam-LT} \\
&j \gets \Argmin_{i\in \{1,\dots,K\}}L_B(\theta_A,\theta_B^i), \tag{6b-i} \label{eq:kbeam-LB-argMAX} \\
&\theta_A \gets \theta_A + \alpha_A \nabla_{\theta_A} L_B(\theta_A,\theta_B^j), \tag{6b-ii} \label{eq:kbeam-LB-MAX} \\
&\theta_B^i \gets \theta_B^i - \alpha_B \nabla_{\theta_B^i} L_B(\theta_A, \theta_B^i),\ \forall i \in \{1,\dots,K\}. \tag{6c} \label{eq:kbeam-LB-MIN}
\end{align}
\end{subequations}

We denote this method as \textbf{Ours-$K$-Beam} in the following parts and give the details in Algorithm~\ref{alg:kbeam}, where $d\_iter$ is the number of iterations used in the step of maximizing $L_B$. 

\begin{algorithm}
\SetAlgoLined
Initialize $\theta_A$, $\theta_{T}$ and $\{\theta_B^i\}_{i=1}^K$\;
\For{$t_0 \gets 1$ \KwTo max\_iter}
{
    \textbf{/*$L_T$ step:*/} \\
    \While{$Acc(f_T(f_A(X^v)),Y_T^v)$ $\leq th_T$ }
    {
        Update $\theta_T$, $\theta_A$ using Eq.~(\ref{eq:kbeam-LT})
    }
    \textbf{/*$L_B$ \emph{Max} step:*/} \\
    Update $j$ using Eq.~(\ref{eq:kbeam-LB-argMAX}) \\
    \For{$t_1\gets1$ \KwTo d\_iter}
    {
        Update $\theta_A$ using Eq.~(\ref{eq:kbeam-LB-MAX})
    }
    \textbf{/*$L_B$ \emph{Min} step:*/} \\
    \For{$i\gets1$ \KwTo K}
    {
        \While{$Acc(f_B^i(f_A(X^t)),Y_B^t)$ $\leq th_B$}
        {
            Update $\theta_B^i$ using Eq.~(\ref{eq:kbeam-LB-MIN})
        }
    }
}
\caption{Ours-$K$-Beam Algorithm}
\label{alg:kbeam}
\end{algorithm}

\subsubsection{Maximize entropy} \label{sec:Nentropy}

We empirically find that minimizing negative cross-entropy $J_B$, which is a \emph{concave} function, causes numerical instabilities in Eq.~(\ref{eq:GRL-D}). 
So, we replace $J_B$ with $-H_B$, the negative entropy of $f_B(f_A(X))$, which is a \emph{convex} function\footnote{{This point discusses the convexity or concavity of different loss functions when viewing them as the outermost function in the composite function. Both loss functions are neither convex nor concave \wrt~model weights.}}:
\begin{align*}
    H_B(f_B(f_A(X))) \triangleq H(f_B(f_A(X))),
\end{align*}
where $H(\cdot)$ is the entropy function.
Minimizing $-H_B$ is equivalent to maximizing entropy, which will encourage ``uncertain'' predictions. 
We replace $J_B$ in Eq.~(\ref{eq:GRL-D}) by $-H_B$, abbreviate  $H_B(f_B(f_A(X)))$ as $H_B(\theta_A, \theta_B)$, and propose the following new update scheme:
\begin{subequations}
\begin{align} % & indicates where to align
&\theta_A \gets \theta_A - \alpha_A \nabla_{\theta_A}(L_T(\theta_A, \theta_T) - \gamma H_B(\theta_A, \theta_B)), \label{eq:ME-D} \\
&\theta_T,\theta_A \gets \theta_T, \theta_A - \alpha_T \nabla_{\theta_T, \theta_A} L_T(\theta_A,\theta_T), \label{eq:ME-T} \\
&\theta_B \gets \theta_B - \alpha_B \nabla_{\theta_B} L_B(\theta_A, \theta_B), \label{eq:ME-B}
\end{align}
\end{subequations}
where $L_T$ and $L_B$ are still cross-entropy loss functions as in Eq.~(\ref{eq:GRL-loss}). 
Unlike in Eq. (\ref{eq:GRL-T}), where we only update $\theta_T$ when minimizing $L_T$, we train $\theta_T$ and $\theta_A$ in an end-to-end manner as shown in Eq.~(\ref{eq:ME-T}), since we find it achieves better performance in practice.

We denote this method as \textbf{Ours-Entropy} in the following parts and give the details in Algorithm~\ref{alg:ME}.

\begin{algorithm}
\SetAlgoLined
Initialize $\theta_A$, $\theta_{T}$ and $\theta_{B}$\;
\For{$t_0 \gets 1$ \KwTo max\_iter}
{
    Update $\theta_A$ using Eq.~(\ref{eq:ME-D}) \\
    \While{$Acc(f_T(f_A(X^v)),Y_T^v)$ $\leq th_T$ }
    {
        Update $\theta_T$, $\theta_A$ using Eq.~(\ref{eq:ME-T})
    }
    \While{$Acc(f_B(f_A(X^t)),Y_B^t)$ $\leq th_B$}
    {
        Update $\theta_{B}$ using Eq.~(\ref{eq:ME-B})
    }
}
\caption{Ours-Entropy Algorithm}
 \label{alg:ME}
\end{algorithm}

\subsection{Addressing the Dataset Challenge by Cross-Dataset Training and Evaluation: An Initial Attempt}\label{sec:cross-dataset}
An ideal dataset to train and evaluate our framework would be a set of human action videos with both action labels and privacy attributes provided.
On the SBU dataset, we can use the actor pair as a simple privacy attribute. But when we want to evaluate our method on more complex privacy attributes, we run into the dataset challenge: in the literature, no existing datasets have both human action labels and privacy attributes provided on the same videos.

% \noindent \textbf{Transferability of Privacy Attributes Across Datasets} 
Given the observation that a privacy attributes predictor trained on VISPR can correctly identify privacy attributes occurring in UCF101 and HMDB51 videos (examples in the Appendix C), we hypothesize that the privacy attributes have good ``transferability'' across  UCF101/HMDB51 and VISPR. Therefore, we can use a privacy prediction model trained on VISPR to assess the privacy leak risk on UCF101/HMDB51.

% \noindent \textbf{Cross-Dataset Training and Evaluation: An Initial Solution} 
In view of that, we propose to use cross-dataset training and evaluation as a workaround method.
In brief, we train action recognition (target utility task) on human action datasets, such as UCF101\cite{soomro2012ucf101} and HMDB51\cite{kuehne2011hmdb}, and train privacy protection (budget task) on visual privacy dataset VISPR\cite{orekondy2017towards}, while letting the two interact via their shared component - the learned anonymization model. 
More specifically, during training, we have two pipelines: one is $f_A$ and $f_T$ trained on UCF101 or HMDB51 for action recognition; the other is $f_A$ and $f_B$ trained on VISPR to suppress multiple privacy attribute prediction. The two pipelines share the same parameters for $f_A$. 
During the evaluation, we evaluate model utility (\ie, action recognition) on the testing set of UCF101 or HMDB51 and privacy protection performance on the testing set of VISPR.
Such cross-dataset training and evaluation shed new possibilities on training privacy-preserving recognition models, even under the practical shortages of datasets that have been annotated for both tasks. 
Notably, ``cross-dataset training'' and ``cross-dataset testing (or evaluation)'' are two independent strategies used in this paper; they can be used either together or separately. Details of our three experiments (SBU, UCF-101, and HMDB51) are explained as follows:
\begin{itemize}[leftmargin=*]
    \item SBU (Section 4.1): we train and evaluate our framework on the same video set by considering actor identity as a simple privacy attribute. \emph{Neither cross-training nor cross-evaluation is involved}.
    \item UCF101 (Section 4.2): we perform \emph{both cross-training and cross-evaluation}, on UCF-101 + VISPR. Such a method provides an alternative to flexibly train and test privacy-preserving video recognition for different utility/privacy combinations, without annotating specific datasets. 
    \item HMDB51 (Section 5.5), we use cross-training on HMDB51 + VISPR datasets similarly to the UCF-101 experiment; but for testing, we evaluate both utility and privacy performance on the same, newly-annotated PA-HMDB51 testing set. Therefore, it involves \emph{cross-training, but no cross-evaluation}. 
\end{itemize}

Beyond the above initial attempt, we further construct a new dataset dedicated to the privacy-preserving action recognition task, which will be presented in Section~\ref{sec:dataset}.

\subsection{Addressing the $\forall$ Challenge by Privacy Budget Model Restarting and Ensemble} \label{sec:address-any}
To improve the generalization ability of learned $f_A$ over all possible $f_B \in \mathcal{P}$ (\ie, privacy cannot be reliably predicted by any model), we hereby discuss two simple and easy-to-implement options. Other more sophisticated model re-sampling or model search approaches, such as~\cite{Zoph2018LearningTA}, will be explored in future work.

\subsubsection{Privacy Budget Model Restarting}
\vspace{+0.25em}
\noindent \textbf{Motivation.}
The max step over $J_B(f_B(f_A(X)), Y_B)$ in Eq.~(\ref{eq:loss-max}) leads to the optimizer being stuck in bad local solutions (similar to ``mode collapse'' in GANs), that will hurdle the entire minimax optimization. Model restarting provides a mechanism to ``bypass'' the bad solution when it occurs, thus enabling the minimax optimizer to explore better solution.

\vspace{+0.25em}
\noindent \textbf{Approach.} 
At certain point of training (\eg, when the privacy budget $L_B(f_B(f_A(X)),Y_B)$ stops decreasing any further), we re-initialize $f_B$ with random weights. Such a random restarting aims to avoid trivial overfitting between $f_B$ and $f_A$ (\ie, $f_A$ is only specialized at confusing the current $f_B$), without requiring more parameters. We then start to train the new model $f_B$ to be a strong competitor, \wrt the current $f_A(X)$: specifically, we freeze the training of $f_A$ and $f_T$, and change to minimizing $L_B(f_B(f_A(X)),Y_B)$, until the new $f_B$ has been trained from scratch to become a strong privacy prediction model over current $f_A(X)$. We then resume adversarial training by unfreezing $f_A$ and $f_T$, as well as switching the loss for $f_B$ back to the adversarial loss (negative entropy or negative cross-entropy). Such a random restarting can repeat multiple times.

\subsubsection{Privacy Budget Model Ensemble}
\vspace{+0.25em}
\noindent \textbf{Motivation.} 
Ideally in Eq.~(\ref{eq:loss-max}) we should maximize the error over the ``current strongest possible'' attacker $f_A$ from $\mathcal{P}$ (a large and continuous $f_B$ family), over which searching/sampling is impractical. Therefore we propose a privacy budget model ensemble as an approximation strategy, where we approximate the continuous $\mathcal{P}$ with a discrete set of $M$ sample functions. Such a strategy is empirically verified in Section 4 and 5 to address the critical ``$\forall$ Challenge'' in privacy protection, \ie, enhancing the defense against unseen attacker models (compared to the clear ``attacker overfitting'' phenomenon when sticking to one $f_A$ during training). 

\vspace{+0.25em}
\noindent \textbf{Approach.} 
Given the budget model ensemble $\bar{\mathcal{P}}_t \triangleq \{f_B^i\}_{i=1}^M$, where $M$ is the number of $f_B$s in the ensemble during training, we turn to minimize the following discretized surrogate of Eq.~(\ref{eq:loss-sup}):
\begin{align} 
\begin{split} \label{eq:loss-ensemble}
f_A^*, f_T^* = \Argmin_{f_A,f_T}&[L_T(f_T(f_A(X)),Y_T) + \\
\gamma & \Max_{f_B^i \in \bar{\mathcal{P}}_t }J_B(f_B^i(f_A(X)),Y_B)].
\end{split}
\end{align}
The previous basic framework is a special case of Eq.~(\ref{eq:loss-ensemble}) with $M = 1$. The ensemble strategy can be easily incorporated with restarting. 

\subsubsection{Incorporate Budget Model Restarting and Budget Model Ensemble with Ours-Entropy}
Budget Model Restarting and Budget Model Ensemble can be easily incorporated with all three optimization schemes described in Section~\ref{sec:algos}. We take Ours-Entropy as an example here.
When model ensemble is used, we abbreviate $L_B(f_B^i(f_A(X)), Y_B)$ and $H_B(f_B^i(f_A(X)))$ as $L_B(\theta_A, \theta_B^i)$ and $H_B(\theta_A, \theta_B^i)$ respectively.
The new parameter updating scheme is:
\begin{subequations} \label{eq:GRL-loss-M}
\begin{flalign} % & indicates where to align
& \theta_A \gets \theta_A - \alpha_A \nabla_{\theta_A} (L_T + \gamma \Max_{\theta_B^i \in \bar{\mathcal{P}}_t} -H_B(\theta_A, \theta_B^i)), \hspace{-2pt} \label{eq:ME-M-D} \\
& \theta_A, \theta_T \gets \theta_A, \theta_T - \alpha_T \nabla_{(\theta_A,\theta_T)} L_T(\theta_A, \theta_T), \label{eq:ME-M-T} \\
& \theta_B^i \gets \theta_B^i - \alpha_B \nabla_{\theta_B^i} L_B(\theta_A, \theta_B^i),\ \forall i \in \{1,\dots,M\}. \label{eq:ME-M-B}
\end{flalign}
\end{subequations}

That's to say, we only suppress the model $f_B^i$ with the largest privacy leakage $-H_B$, \ie, the ``most confident'' one about its current privacy prediction, when updating the anonymization model $f_A$. 
But we still update all $M$ budget models on the budget task. 
The formal description of Ours-Entropy with model restarting and ensemble is given in Algorithm~\ref{alg:ME2}, where $\{\theta_B^i\}_{i=1}^M$ is reinitialized every $rstrt\_iter$ iterations. Likewise, GRL and Our-$K$-Beam can also be incorporated with restarting and ensemble, whoses details are shown in Appendix A. 
\begin{algorithm} 
\SetAlgoLined
Initialize $\theta_A$, $\theta_{T}$ and $\{\theta_B^i\}_{i=1}^M$\;
\For{$t_0 \gets 1$ \KwTo max\_iter}
{
    \If{$t\equiv 0 \pmod{rstrt\_iter}$}
    {
        Reinitialize $\{\theta_B^i\}_{i=1}^M$ \\
    }
    Update $\theta_A$ using Eq.~(\ref{eq:ME-M-D}) \\
    \While{$Acc(f_T(f_A(X^v)),Y_T^v)$ $\leq th_T$ }
    {
        Update $\theta_T, \theta_A$ using Eq.~(\ref{eq:ME-M-T})
    }
    \For{$i\gets 1$ \KwTo $M$}
    {
        \While{$Acc(f_B^i(f_A(X^t)),Y_B^t)$ $\leq th_B$}
        {
            Update $\theta_B^i$ using Eq.~(\ref{eq:ME-M-B})
        }
    }
}
\caption{Ours-Entropy Algorithm (with Model Restarting and Model Ensemble)}
\label{alg:ME2}
\end{algorithm}

\subsection{Two-Step Evaluation Protocol} \label{sec:two-step}
% The evaluation protocol for privacy-preserving visual recognition is more intricate than classical visual recognition tasks in order to balance between the two task models. 
The solution to Eq.~(\ref{eq:loss-sup}) gives an anonymization model $f_A^*$ and a target utility task model $f_T^*$.  We need to evaluate $f_A^*$ and $f_T^*$ on the trade-off they have achieved between target task utility and privacy protection in two steps: (1) whether the learned target utility task model maintains satisfactory performance on anonymized videos; (2) whether the performance of an \textit{arbitrary} privacy prediction model on anonymized videos will deteriorate.

Suppose we have a training dataset $X^t$ with target and budget task ground truth labels $Y_T^t$ and $Y_B^t$, and an evaluation dataset $X^e$ with target and budget task ground truth labels $Y_T^e$ and $Y_B^e$. In the first step, when evaluating the target task utility, we should follow the traditional routine: compare $f_T^*(f_A^*(X^e))$ with $Y_T^e$ to get the evaluation accuracy on the target utility task, denoted as $A_T$, which we expect to be as high as possible.
In the second step, when evaluating the privacy protection, it is insufficient if we only observe that the learned $f_A^*$ and $f_B^*$ lead to poor classification accuracy on $X^e$, because of the $\forall$ challenge:
the attacker can select \textit{any privacy budget model} to steal private information from anonymized videos $f_A^*(X^e)$.
To empirically verify that $f_A^*$ prohibits reliable privacy prediction for other possible budget models, we propose a novel procedure:
\begin{itemize}[leftmargin=*]
    \item We randomly re-sample $N$ privacy budget prediction models $\bar{\mathcal{P}}_e \triangleq \{f_B^i\}_{i=1}^N$ from $\mathcal{P}$ for evaluation. Note that these $N$ models used in evaluation $\bar{\mathcal{P}}_e$ have no overlap with the $M$ privacy budget model ensemble $\bar{\mathcal{P}}_t$ used in training, \ie, $\bar{\mathcal{P}}_e \cap \bar{\mathcal{P}}_t=\emptyset$.
    \item We train these $N$ models $\bar{\mathcal{P}}_e$ on anonymized training videos $f_A^*(X^t)$ to make correct predictions on private information, \ie, $\min_{f_B^i} L_B(f_B^i(f_A^*(X^t)),Y_B^t), \forall f_B^i \in \bar{\mathcal{P}}_e$. Note that $f_A^*$ is fixed during this training procedure.
    \item After that, we apply each $f_B^i$ on anonymized evaluation videos $f_A^*(X^e)$ and compare the outputs $f_B^i(f_A^*(X^e))$ with $Y_B^e$ to get privacy budget accuracy of the $i$-th budget model, \ie, $Acc(f_B^i(f_A^*(X^e)), Y_B^e)$.
    \item We select the highest accuracy among all $N$ privacy budget models and use it as the final privacy budget accuracy $A_B^N$, which we expect to be as low as possible. Specifically, we have
    \begin{equation}
        A_B^N = \Max_{f_B^i \in \bar{\mathcal{P}}_e} Acc(f_B^i(f_A^*(X^e)), Y_B^e).
    \end{equation}
\end{itemize}
\section{Simulation Experiments}\label{sec:experiments}
We show the effectiveness of our framework on \emph{privacy-preserving action recognition} on existing datasets.

\vspace{+0.25em}
\noindent \textbf{Overview of Experiment Settings.} 
The target utility task is human action recognition, since it is a highly demanded feature in smart home and smart workplace applications. 
Experiments are carried out on three widely used human action recognition datasets: SBU Kinect Interaction Dataset\cite{Yun2012TwopersonID}, UCF101\cite{soomro2012ucf101} and HMDB51\cite{kuehne2011hmdb}.
The privacy budget task varies in different settings. In the SBU dataset experiments, the privacy budget is to prevent the videos from leaking human identity information. In the experiments on UCF101 and HMDB51, the privacy budget is to protect visual privacy attributes as defined in~\cite{orekondy2017towards}. 
We emphasize that the general framework proposed in Section~\ref{sec:framework} can be used for a large variety of target utility tasks and privacy budget task combinations, not only limited to the aforementioned settings.

Following the notations in Section~\ref{sec:framework}, on all the video action recognition datasets including SBU, UCF101 and HMDB51, we set $W=112$, $H=112$, $C=3$, and $T=16$ (C3D's required temporal length and spatial resolution).
Note that the original resolution for SBU, UCF101 and HDMB51 are $640 \times 480$, $320 \times 240$ and $320 \times 240$, respectively. We downsample video frames to resolution $160 \times 120$. To reduce the spatial resolution to $112 \times 112$, we use random-crop and center-crop in training and evaluation, respectively.

\vspace{+0.25em}
\noindent \textbf{Baseline Approaches.} 
We consistently use two groups of approaches as baselines across the three action recognition datasets. 
% These two groups of baselines are \emph{naive downsample at different rates} and \emph{empirical obfuscations}. The group of \emph{naive downsample at different rates} chooses rates from $\{1,2,4,8,16\}$, where $1$ stands for no down-sampling.
These two groups of baselines are \emph{naive downsamples} and \emph{empirical obfuscations}. The group of \emph{naive downsamples} chooses downsample rates from $\{1,2,4,8,16\}$, where $1$ stands for no down-sampling.
The group of \emph{empirical obfuscations} includes approaches selected from different combinations in \{box, segmentation\} $\times$ \{blurring, blackening\} $\times$ \{face, human body\}. Details are listed below:
\begin{itemize}[leftmargin=*]
\item \emph{Naive Downsamle}: Spatially downsample each frame.
\item \emph{Box-Black-Face}: Boxing and blackening faces.
\item \emph{Box-Black-Body}: Boxing and blackening bodies.
\item \emph{Seg-Black-Face}: Segmenting and blackening faces.
\item \emph{Seg-Black-Body}: Segmenting and blackening bodies.
\item \emph{Box-Blur-Face}: Boxing and blurring faces.
\item \emph{Box-Blur-Body}: Boxing and blurring bodies.
\item \emph{Seg-Blur-Face}: Segmenting and blurring faces.
\item \emph{Seg-Blur-Body}: Segmenting and blurring bodies.
\end{itemize}

\vspace{+0.25em}
\noindent \textbf{Our Proposed Approaches.} 
The previous two groups of baselines are compared with our proposed three approaches:
\begin{itemize}[leftmargin=*]
\item \emph{GRL}: as described in Section~\ref{sec:grl}.
\item \emph{Ours-$K$-Beam}: as described in Section~\ref{sec:kbeam}. We have tried $K=1,2,4,8$.
\item \emph{Ours-Entropy}: as described in Section~\ref{sec:Nentropy}. In the privacy budget model ensemble $\bar{\mathcal{P}}_t$, the $M$ models are chosen from MobileNet-V2~\cite{sandler2018mobilenetv2} family with different width multipliers. We have tried $M=1,2,4,8$.
\end{itemize}
All three approaches are evaluated with and without privacy budget model restarting.

\vspace{+0.25em}
\noindent \textbf{Evaluation.} 
In the two-step evaluation (as described in Section~\ref{sec:two-step}), we have used $N=10$ different state-of-the-art classification networks, namely ResNet-V1-\{50,101\}~\cite{He_2016}, ResNet-V2-\{50,101\}~\cite{he2016identity}, Inception-V1~\cite{szegedy2015going}, Inception-V2~\cite{szegedy2016rethinking}, and MobileNet-V1-\{0.25,0.5,0.75,1\}~\cite{howard2017mobilenets}, as $\bar{\mathcal{P}}_e$. Note that $\bar{\mathcal{P}}_e \cap \bar{\mathcal{P}}_t = \emptyset$. All detailed numerical results reported in following sections can be found in Appendix B.

% SBU:
\subsection{Identity-Preserving Action Recognition on SBU: Single-Dataset Training}\label{sec:exp-sbu}
We compare our proposed approaches with the groups of baseline approaches to show our methods' significant superiority in balancing privacy protection and model utility. 
We use three different optimization schemes described in Section~\ref{sec:algos} on our framework and empirically show all three largely outperform the baseline methods.
We also show that adding the model ensemble and model restarting, as described in Section~\ref{sec:address-any}, to the optimization procedure can further improve the performance of our method.

\subsubsection{Experiment Setting}
SBU Kinect Interaction Dataset\cite{Yun2012TwopersonID} is a two-person interaction dataset for video-based action recognition. 7 participants performed actions, and the dataset is composed of 21 sets. Each set uses different pairs of actors to perform all 8 interactions. 
However, some sets use the same two actors but with different actors acting and reacting. For example, in set 1, actor 1 is acting, and actor 2 is reacting; in set 4, actor 2 is acting, and actor 1 is reacting. These two sets have the same actors, so we combine them as one class to better fit our experimental setting.
In this way, we combine all sets with the same actors and finally get 13 different actor pairs.
This dataset's target utility task is action recognition, which could be taken as a classification task with 8 different classes. The privacy budget task is to recognize the actor pairs of the videos, which could be taken as a classification task with 13 different classes.

\subsubsection{Implementation Details}
In Algorithms~\ref{alg:GRL}-\ref{alg:ME}, we set step sizes $\alpha_T=10^{-5}$, $\alpha_B=10^{-2}$, $\alpha_A=10^{-4}$, accuracy thresholds $th_T=85\%$, $th_B=99\%$ and $max\_iter=800$.
In Algorithm~\ref{alg:kbeam}, we set $d\_iter$ to be $30$. 
In Algorithm~\ref{alg:ME2}, we set $rstrt\_iter$ to be $100$.
Other hyper-parameters of Algorithm~\ref{alg:ME2} are identical with those in Algorithm~\ref{alg:ME}.
We set $\gamma$=$2$ in Eq.~(\ref{eq:loss-max}) and use Adam optimizer~\cite{Kingma2015AdamAM} to update all parameters.

\begin{figure}[!t]
\centering
\includegraphics[width=\columnwidth]{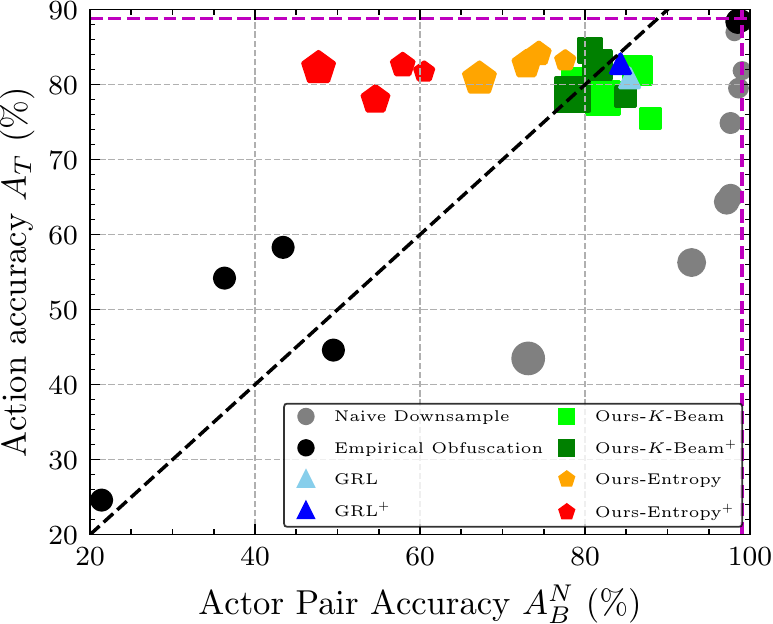}
\vspace{-1.5em}
\caption{The trade-off between privacy budget and action utility on SBU dataset. For \emph{Naive Downsample} method, a larger marker means a larger adopted down-sampling rate. For \emph{Ours-$K$-Beam} method, a larger marker means a larger $K$ (number of beams) in Algorithm~\ref{alg:kbeam}. For \emph{Ours-Entropy} and \emph{Ours-Entropy (restarting)}, a larger marker means a larger $M$ (number of ensemble models) in Algorithm~\ref{alg:ME2}.
Methods with ``+'' superscript are incorporated with model restarting.
Vertical and horizontal purple dashed lines indicate $A_B^N$ and $A_T$ on the original non-anonymized videos, respectively.
The black dashed line indicates where $A_B^N=A_T$.
Detailed experimental settings and numerical results for each method can be found in Appendix B.
}
\label{fig:SBU1}
\end{figure}

\subsubsection{Results and Analyses}
We present the experimental results of our proposed methods and other baseline methods in Figure~\ref{fig:SBU1}, which shows the trade-off between the action recognition accuracy $A_T$, and the actor pair recognition accuracy $A_B^N$. 
In order to interpret this figure, we should note that a desirable trade-off should incur maximal target accuracy $A_T$ (y-axis) and minimal privacy budget accuracy $A_B^N$ (x-axis). Therefore, a point closer to the top-left corner represents an anonymization model $f_A^*$ with more desirable performance. 
The magenta dotted line suggests the target accuracy $A_T$ on original unprotected videos. This can be roughly considered as the $A_T$ upper bound for all privacy protection methods, under the assumption that $f_A^*$ will unavoidably filter out some useful information for the target utility task.

As we can see, Ours-$K$-Beams, Ours-Entropy, and GRL all largely outperform the two groups of naive baselines. 
\{box, segmentation\} $\times$ \{blurring, blackening\} $\times$ \{face\} and naive downsample with a low rate (\eg, $2$ and $4$) can lead to decent action accuracy, but the privacy budget accuracy $A_B^N$ is still very high, meaning these methods fail to protect privacy. 
On the other hand, \{box, segmentation\} $\times$ \{blurring, blackening\} $\times$ \{body\} and naive downsample with a high rate (\eg, $8$ and $16$) can effectively suppress $A_B^N$ to a low level, but $A_T$ also suffers a huge negative impact, which means the anonymized videos are of little practical utility.  
Our methods, in contrast, achieve a great balance between utility and privacy protection. 
Ours-Entropy can decrease $A_B^N$ by around $30\%$ with nearly no harm on $A_T$. 

\vspace{+0.25em}
\noindent\textbf{Comparison of three methods.}
$K$-Beam is a state-of-the-art minimax optimization problem, and we apply it to solve a sub-problem (\ie, Eq.~(\ref{eq:kbeam-minimax})) of our more complex three-party competition game. Unfortunately, we empirically find the $K$-Beam algorithm becomes more unstable when we introduce a new competing party to the minimax optimization problem. 
GRL is originally proposed for domain adaptation. On our new visual privacy protection task, we find it unstable and sensitive to model initialization.  
By replacing the \emph{concave} negative cross-entropy loss function with the \emph{convex} negative entropy, Ours-Entropy empirically stabilizes the optimization and gives the best performance among all three methods.

The results also show the effectiveness of model restarting and model ensemble: model restarting can further suppress $A_B^N$ with little harm on $A_T$, and model ensemble with larger $M$ can improve the trade-off even more. 
% Due to its effectiveness and low complexity, we use Ours-Entropy with restarting as the default option in our framework, unless otherwise noted.

% UCF101:
\subsection{Action Recognition on UCF101 with Multiple Privacy Attributes Protected: Cross-Dataset Training and Evaluation}\label{sec:exp-ucf101}

\subsubsection{Experiment Setting}
UCF101 is an action recognition dataset with 13,320 real-life human action videos collected from YouTube. It contains videos of 101 different actions.
We use the official train-test split for this dataset.

The target utility task $\mathcal T$ is to do human action recognition on UCF101, which can be taken as a video classification task with 101 classes.

VISPR is a dataset with 22,167 images annotated with 68 privacy attributes, \eg, semi-nudity, hobbies, face, race, gender, skin color, and so on.
Each attribute of an image is labeled as ``present'' or ``non-present'' depending on whether the specific privacy attribute information is contained in the image.
Among the 68 attributes, there are 7 attributes that frequently appear in both UCF101 datasets and the smart home videos. Therefore we select these 7 attributes for protection in our experiments. These 7 attributes are semi-nudity, occupation, hobbies, sports, personal relationship, social relationship, and safe.
The privacy budget task $\mathcal B$ is to predict privacy attributes on the VISPR dataset, which can be taken as a multi-label image classification task (7 labels, each is a binary classification task). 
% We use class-based mean average precision (cMAP), whose definition can be referred to~\cite{orekondy2017towards}, to measure the performance of this privacy budget task.
We adopt the class-based mean average precision (cMAP)~\cite{orekondy2017towards} as $A_B^N$ to measure the performance of the privacy budget task.
The official train-test split is used on the VISPR dataset.

\begin{figure}[h]
\centering
\includegraphics[width=\columnwidth]{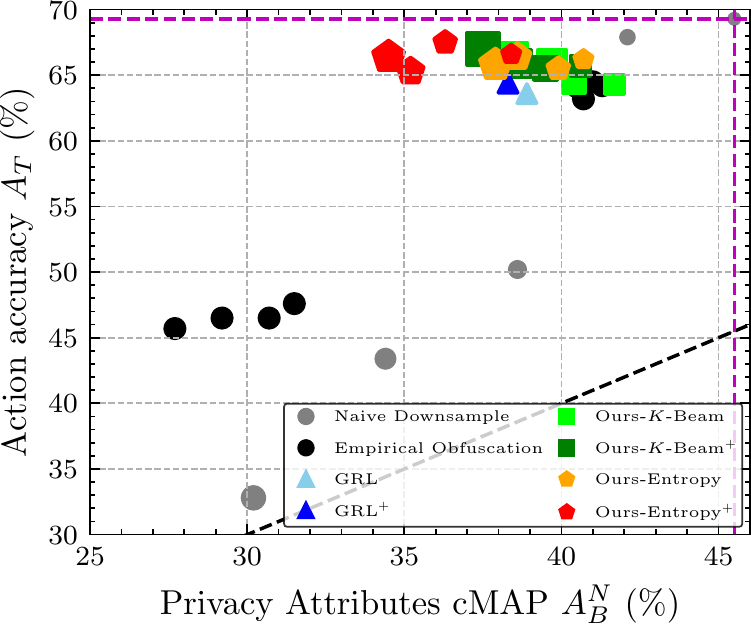}
\vspace{-1.5em}
\caption{The trade-off between privacy budget and action utility on UCF-101/VISPR Dataset. For \emph{Naive Downsample} method, a larger marker means a larger down sampling rate is adopted. For \emph{Ours-$K$-Beam} method, a larger marker means a larger $K$ (number of beams) in Algorithm~\ref{alg:kbeam}. For \emph{Ours-Entropy} and \emph{Ours-Entropy (restarting)}, a larger marker means a larger $M$ (number of ensemble models) in Algorithm~\ref{alg:ME2}.
Methods with ``+'' superscript are incorporated with model restarting.
Vertical and horizontal purple dashed lines indicate $A_B^N$ and $A_T$ on the original non-anonymized videos, respectively.
The black dashed line indicates where $A_B^N=A_T$.
Detailed experimental settings and numerical results for each method can be found in Appendix B.
}
\label{fig:UCF101}
\end{figure}

\subsubsection{Implementation Details}

In Algorithm~\ref{alg:ME2}, we set step sizes $\alpha_T=10^{-5}$, $\alpha_B=10^{-2}$, $\alpha_A=10^{-4}$, accuracy thresholds $th_T=70\%$, $th_B=99\%$, $max\_iter=800$ and $rstrt\_iter=100$.
We set $\gamma=0.5$ in Eq.~(\ref{eq:loss-max}) and use Adam optimizer to update all parameters. 
Values of $K$ and $M$ are identical to those in SBU experiments.

\subsubsection{Results and Analyses}
We present the experimental results in Figure~\ref{fig:UCF101}. 
All naive downsample and empirical obfuscation methods cause $A_T$ to drop dramatically while $A_B^N$ only drops a little bit, which means the utility of videos is greatly reduced while the private information is hardly filtered out. 
In contrast, with the help of model restarting and model ensemble, Ours-Entropy can decrease $A_B^N$ by $7\%$ while keeping $A_T$ as high as that on the original raw videos, meaning the privacy is protected at almost no cost on the utility.
Hence, Ours-Entropy outperforms all naive downsample and empirical obfuscation baselines in this experiment.
It also shows an advantage over GRL and Ours-$K$-Beam.

\subsection{Anonymized Video Visualization}
\begin{figure}[!t]
    \captionsetup[subfigure]{labelformat=empty,justification=centering,farskip=2pt,captionskip=1pt}
    \centering
    \subfloat
    {
       \includegraphics[width=0.23\linewidth]{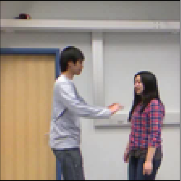}
    }
    \subfloat
    {
       \includegraphics[width=0.23\linewidth]{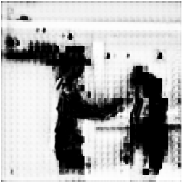}
    }
    \subfloat
    {
       \includegraphics[width=0.23\linewidth]{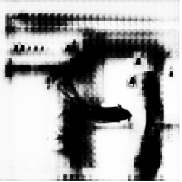}
    }
    \subfloat
    {
       \includegraphics[width=0.23\linewidth]{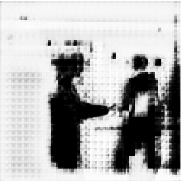}
    }
    \\
    \subfloat
    {
       \includegraphics[width=0.23\linewidth]{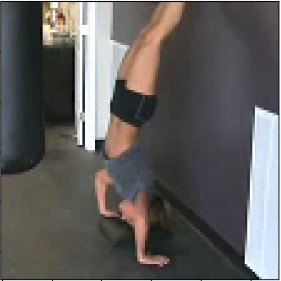}
    }
    \subfloat
    {
       \includegraphics[width=0.23\linewidth]{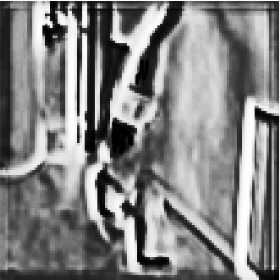}
    }
    \subfloat
    {
       \includegraphics[width=0.23\linewidth]{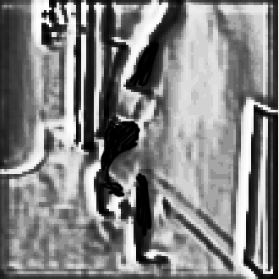}
    }
    \subfloat
    {
       \includegraphics[width=0.23\linewidth]{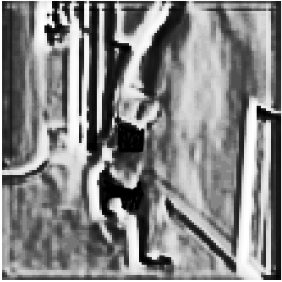}
    }
    \\
    \subfloat[Original]
    {
        \includegraphics[width=0.23\linewidth]{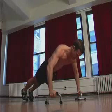}
    }
    \subfloat[M=1]
    {
        \includegraphics[width=0.23\linewidth]{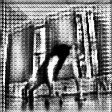}
    } 
    \subfloat[$\text{M=1}^+$]
    {
        \includegraphics[width=0.23\linewidth]{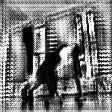}
    } 
    \subfloat[$\text{M=4}^+$]
    {
        \includegraphics[width=0.23\linewidth]{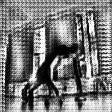}
    }
    \vspace{-0.5em}
    \caption{The center frame of example videos before (column 1) and after (columns 2-4) applying the anonymization transform learned by Ours-Entropy.
    The first row shows a frame from a ``pushing'' video in the SBU dataset; the second row shows a frame from a ``handstand'' video in the UCF101 dataset; the third row shows a frame from a ``push-up'' video in the PA-HMDB51 dataset. Privacy attributes in the last two rows include semi-nudity, face, gender, and skin color.
    Model restarting and ensemble settings are indicated below each anonymized image. $M$ is the number of ensemble models. Methods with a ``+'' superscript are incorporated with model restarting.
    }
    \label{fig:anonymization-visualize}
\end{figure}

We provide the visualization of the anonymized videos on SBU, UCF101, and our new dataset PA-HMDB51 (see Section~\ref{sec:dataset}) in Figure~\ref{fig:anonymization-visualize}. To save space, we only show the center frame of each anonymized video. The visualization shows that the privacy attributes in the anonymized videos are filtered out, but it is still possible to recognize the actions.

% SBU visual figure:
\section{PA-HMDB51: A New Benchmark} \label{sec:dataset}
\subsection{Motivation}
There is no public dataset containing both human action and privacy attribute labels on the same videos in the literature. This poses two challenges. Firstly, the lack of available datasets has increased the difficulty in employing a data-driven joint training method. Secondly, this complication has made it impossible to directly evaluate the trade-off between privacy budget and action utility achieved by a learned anonymization model $f_A^*$.
To solve this problem, we annotate and present the very first human action video dataset with privacy attributes labeled, named \textbf{PA-HMDB51} (\textbf{P}rivacy-\textbf{A}nnotated HMDB51). 
We evaluate our method on this newly built dataset and further demonstrate our method's effectiveness.

\subsection{Selecting and Labeling Privacy Attributes}
A recent work~\cite{orekondy2017towards} has defined 68 privacy attributes which could be disclosed by images. 
However, most of them seldom make any appearance in public human action datasets.
% We carefully select 7 privacy attributes that are most relevant to our smart home settings out of the 68 attributes from~\cite{orekondy2017towards}.
% The $7$ attributes we use are skin color, gender, face (partial), face (complete), nudity, personal relationship, and social circle.
% We further combine those 7 attributes into 5 to better fit the human action videos: combining ``face (partial)'' and ``face (complete)'' into one attribute ``face'' and combining ``personal relationship'' (only intimate relationships such as friends, couples or family members are considered in our setting) and ``social circle'' (\eg, colleagues and classmates) into ``relationship.''
% To this end, we have 5 privacy attributes that widely appear in public human action datasets and are closely relevant to our smart home setting. 
We carefully select 5 privacy attributes that are most relevant to our smart home settings out of the 68 attributes from~\cite{orekondy2017towards}: skin color, gender, face, nudity, and personal relationship (only intimate relationships such as friends, couples or family members are considered in our setting).
The detailed description of each attribute, their possible ground truth values, and their corresponding meanings are listed in Table~\ref{tab:attribute-def}. Some annotated frames in our PA-HMDB51 dataset are shown in Table~\ref{tab:visual-examples} as examples.

Privacy attributes may vary during the video clip. For example, in some frames, we may see a person's full face, while in the next frames, the person may turn around, and his/her face is no longer visible. Therefore, we decide to label all the privacy attributes on each frame~\footnote{A tiny portion of frames in some HMDB51 videos do not contain any person. No privacy attributes are annotated on those frames.}.

The annotation of privacy labels was manually performed by a group of students at the CSE department of Texas A\&M University. Each video was annotated by at least three individuals and then cross-checked. 

\subsection{HMBD51 as the Data Source}
Now that we have defined the 5 privacy attributes, we need to identify a source of human action videos for annotation. There are a number of choices available, such as~\cite{soomro2012ucf101, kuehne2011hmdb, caba2015activitynet,gu2018ava,kay2017kinetics}. We choose HMDB51~\cite{kuehne2011hmdb} to label privacy attributes since it consists of more diverse private information, especially nudity/semi-nudity and relationship.

We provide a per-frame annotation of the selected 5 privacy attributes on 515 videos selected from HMDB51. In this paper, we treat all 515 videos as testing samples\footnote{Labeling per-frame privacy attributes on a video dataset is extremely labor-consuming and subjective (needing individual labeling then cross-checking). As a result, the current size of PA-HMDB51 is limited. So far, we have only used PA-HMDB51 as the testing set, and we seek to annotate more data and hopefully expand PA-HMDB51 for training as future work.}. Our ultimate goal would be to create a larger-scale version of PA-HMDB51 that allows for both training and testing coherently on the same benchmark. For now, we use PA-HMDB51 to facilitate better testing, while still considering cross-dataset training as a rough yet useful option to train privacy-preserving video recognition (before the larger dataset becomes available).

\subsection{Dataset Statistics}
\subsubsection{Action Distribution} \label{sec:action-distribution}
\begin{figure}[!t]
\centering
\includegraphics[width=0.475\textwidth]{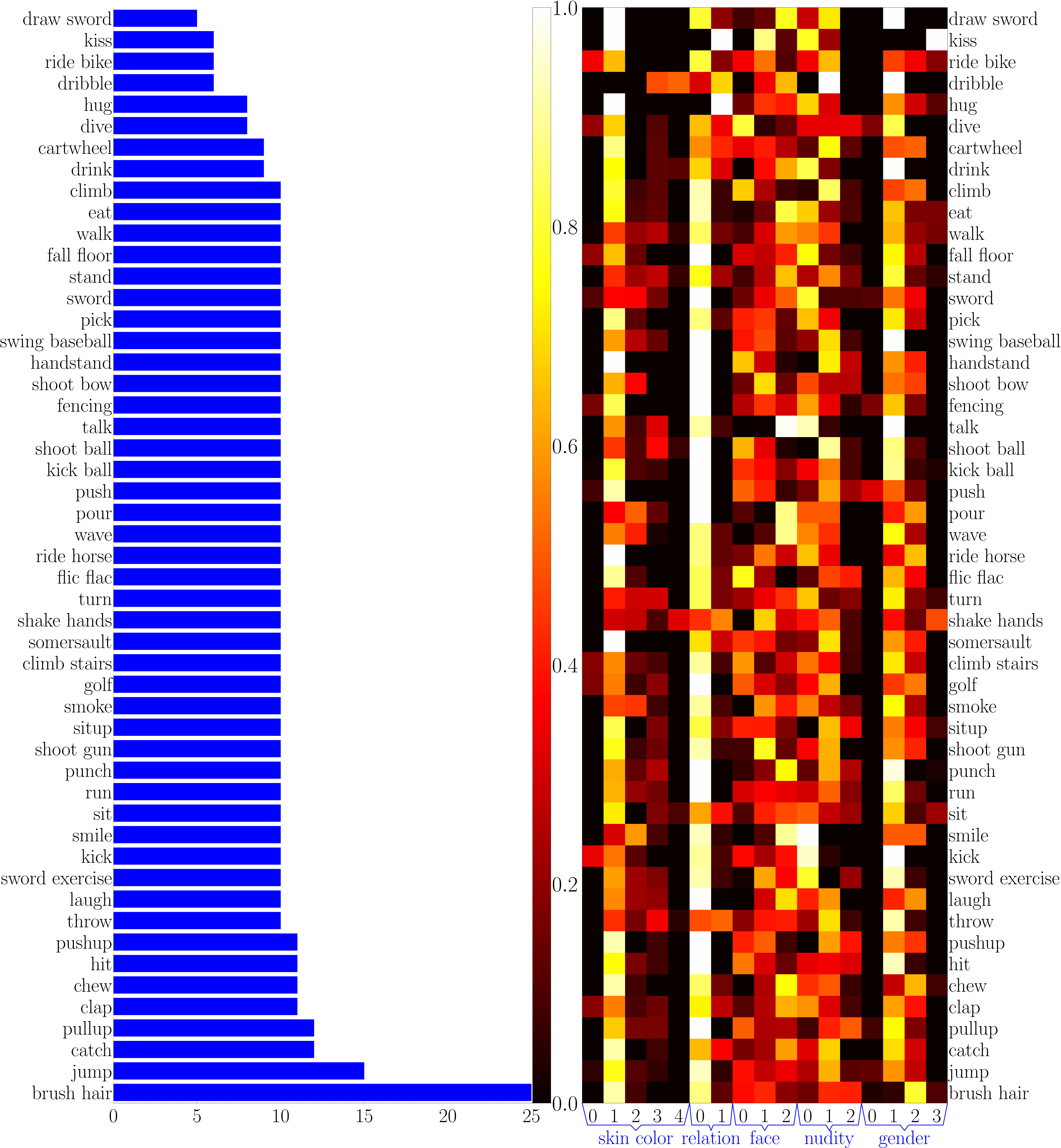}
\vspace{-0.5em}
\caption{\textbf{Left}: action distribution of PA-HMDB51. Each bar shows the number of videos with a certain action. \Eg, the last bar shows there are 25 ``brush hair'' videos in the PA-HMDB51 dataset; \textbf{Right}: action-attribute correlation in the PA-HMDB51 dataset. The $x$-axis are all possible values grouped by bracket for each privacy attribute. The $y$-axis are different action types. The color represents ratio of the number of frames of some action annotated with a specific privacy attribute value \wrt the total number of frames of the action.}
\label{fig:act+act-pa-corl}
\end{figure}

When selecting videos from the HMDB51 dataset, we consider two criteria on action labels. First, the action labels should be balanced. Second (and more implicitly), we select more videos with non-trivial privacy labels.
For example, ``brush hair'' action contains many videos with a ``semi-nudity'' attribute, and ``pull-up'' action contains many videos with a ``partially visible face'' attribute. 
Despite their practical importance, these privacy attributes are relatively less seen in the entire HMDB51 dataset, so we tend to select more videos with these attributes, regardless of their action classes. 
The resultant distribution of action labels is depicted in Figure~\ref{fig:act+act-pa-corl} (left panel), 
showing a relative class balance.

\subsubsection{Privacy Attribute Distribution} \label{sec:pa-distribution}
\begin{figure}[!t]
\centering
\includegraphics[width=0.475\textwidth]{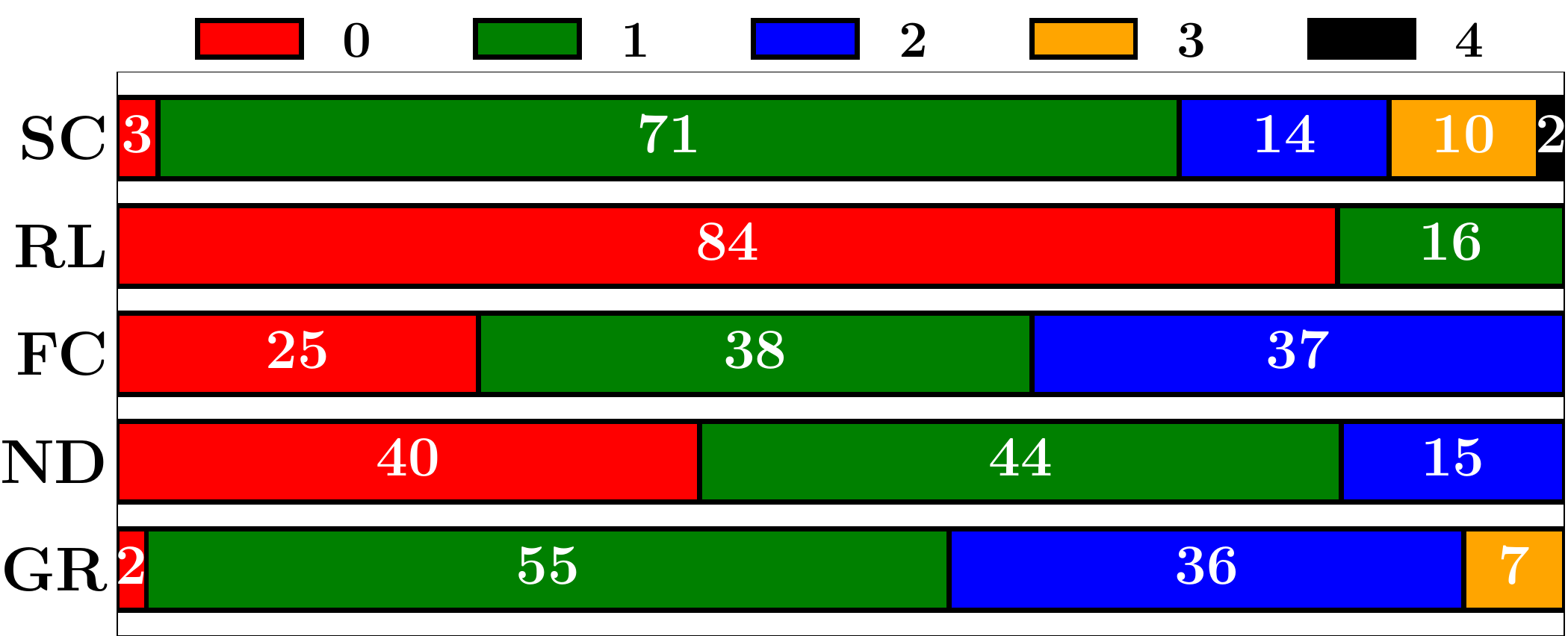}
\vspace{-0.5em}
\caption{Label distribution per privacy attribute in the PA-HMDB51. SC, RL, FC, ND, and GR stand for skin color, relationship, face, nudity, and gender, respectively. The rounded ratio numbers are shown as white text (in $\%$ scale). Definitions of label values $(0,1,2,3,4)$ for each attribute are described in Table~\ref{tab:attribute-def}.}
\label{fig:pa-distribution}
% \vspace{-1em}
\end{figure}

We try to make the label distribution for each privacy attribute as balanced as possible by manually selecting those videos containing uncommon privacy attribute values in original HMDB51 to label. 
For instance, videos with semi-nudity are overall uncommon, so we deliberately select those videos containing semi-nudity into our PA-HMDB51 dataset.
Naturally, people are reluctant to release data that contains privacy concerns to the public, so the privacy attributes are highly unbalanced in any public video datasets.
Although we have used this method to {reduce the data imbalance}, the PA-HMDB51 is still unbalanced.
Frame-level label distributions of all 5 privacy attributes are shown in Figure~\ref{fig:pa-distribution}.

\subsubsection{Action-Attribute Correlation} \label{sec:action-pa}
If there was a strong correlation between a privacy attribute and an action, it would be harder to remove the private information from the videos without much harm to the action recognition task. 
For example, we would expect a high correlation between the attribute ``gender'' and the action ``brush hair'' since this action is carried out much more often by females than by males. 
We show the correlation between privacy attributes and actions in Figure~\ref{fig:act+act-pa-corl} (right panel) and more details in Appendix D.

\begin{table}[!t]
\renewcommand{\arraystretch}{1.3}
\caption{Attribute definition in the PA-HMDB51 dataset}
\vspace{-0.5em}
\label{tab:attribute-def}
\centering
\resizebox{\columnwidth}{!}{
\begin{tabular}{c|l}
\hline
Attribute & \multicolumn{1}{c}{Possible Values \& Meaning}\\
\hline
\hline
\multirow{5}{*}{Skin Color} & \textbf{0}: Skin color of the person(s) is/are \textit{unidentifiable}.\\
& \textbf{1}: Skin color of the person(s) is/are \textit{white}.\\
& \textbf{2}: Skin color of the person(s) is/are \textit{brown/yellow}.\\
& \textbf{3}: Skin color of the person(s) is/are \textit{black}.\\
& \textbf{4}: Persons with different skin colors are \textit{coexisting}.\tablefootnote{
For ``skin color'' and ``gender,'' we allow multiple labels to coexist. For example, if a frame showed a black person's shaking hands with a white person, we would label ``black'' and ``white'' for the ``skin color'' attribute. In the visualization, we use ``coexisting'' to represent the multi-label coexistence and we don't show in detail whether it is ``white and black coexisting'' or ``black and yellow coexisting.''
For the remaining three attributes, we label each attribute using the highest privacy-leakage risk among all persons in the frame. \Eg, given a frame where a group of people are hugging, if there is at least one complete face visible, we would label the ``face'' attribute as ``completely visible.'' 
}\\
\hline
\multirow{3}{*}{Face} & 
\textbf{0}: \textit{Invisible} ($< 10\%$ area is visible).\\
& \textbf{1}: \textit{Partially visible} ($\geq 10\%$ and $\leq 70\%$ area is visible).\\
& \textbf{2}: \textit{Completely visible} ($> 70\%$ area is visible).\\
\hline
\multirow{4}{*}{Gender} & \textbf{0}: The gender(s) of the person(s) is/are \textit{unidentifiable}.\\
& \textbf{1}: The person(s) is/are \textit{male}.\\
& \textbf{2}: The person(s) is/are \textit{female}.\\
& \textbf{3}: Persons with different genders are \textit{coexisting}.\\
\hline
\multirow{3}{*}{Nudity} & \textbf{0}: \textit{No-nudity} with long sleeves and pants.\\
& \textbf{1}: \textit{Partial-nudity} with short sleeves, skirts, or shorts.\\
& \textbf{2}: \textit{Semi-nudity} with half-naked body.\\
\hline
\multirow{2}{*}{Relationship} 
% & \thead[l]{\textbf{0}: \textit{Nonexistent}: No personal relationship is revealed.} \\
% & \thead[l]{\textbf{1}: \textit{Existent}: Personal relationship is revealed.}\\
& {\textbf{0}: Personal relationship is \textit{unidentifiable}.} \\
& {\textbf{1}: Personal relationship is \textit{identifiable}.}\\
\hline
\end{tabular}
}
\end{table}

\begin{table}[!t]
\renewcommand{\arraystretch}{1.3}
\caption{Examples of the annotated frames in the PA-HMDB51 dataset}
\vspace{-0.5em}
\label{tab:visual-examples}
\centering
\resizebox{\columnwidth}{!}{
\begin{tabu}{m{0.18\textwidth} | m{0.05\textwidth} | X[c,m]}
    \hline
    \hfil{Frame} & \hfil{Action} & \hfil{Privacy Attributes} \\
    \hline
    \hline
    \raisebox{-.15\height}{\includegraphics[width=0.18\textwidth, height=0.12\textwidth]{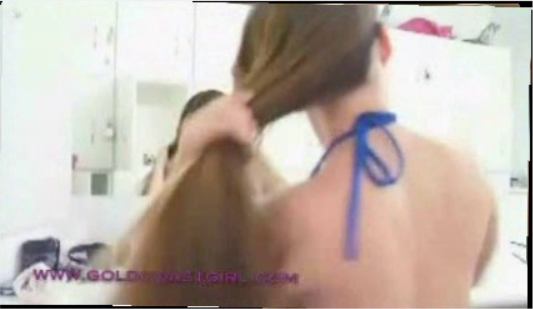}} & 
    Brush hair &
    \begin{center}
    \begin{itemize}[leftmargin=*,label={\tiny\raisebox{0.2ex}{\textbullet}},labelindent=-1.5mm,labelsep=0.5mm]
        \item skin color: white
        \item face: {invisible}
        \item gender: female
        \item nudity: {semi-nudity}
        \item relationship: {unrevealed}
    \end{itemize}
    \end{center}\\ 
    \hline
    \raisebox{-.15\height}{\includegraphics[width=0.18\textwidth, height=0.12\textwidth]{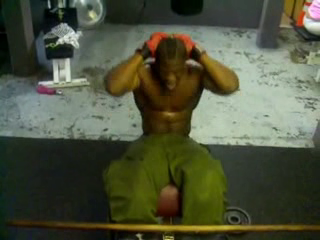}} & 
    Situp & 
    \begin{itemize}[leftmargin=*,label={\tiny\raisebox{0.2ex}{\textbullet}},labelindent=-1.5mm,labelsep=0.5mm]
        \item skin color: black
        \item face: {completely visible}
        \item gender: male
        \item nudity: {semi-nudity}
        \item relationship: {unrevealed}
    \end{itemize} \\ 
    \hline
\end{tabu}
}
\end{table}

% experiments on PA-HMDB51:
\subsection{Benchmark Results on PA-HMDB51: Cross-Dataset Training}\label{sec:exp-hmdb51}

\subsubsection{Experiment Setting}
We train our models using cross-dataset training on HMDB51 and VISPR datasets as we did in Section~\ref{sec:exp-ucf101}, except that we use the 5 attributes defined in Table~\ref{tab:attribute-def} on VISPR instead of the 7 used in Section~\ref{sec:exp-ucf101}. The trained models are directly evaluated on the PA-HMDB51 dataset\footnote{We only use PA-HMDB51 as the testing set so far, since the current size of PA-HMDB51 is limited for training.} for both target utility task $\mathcal{T}$ and privacy budget task $\mathcal{B}$, without any re-training or adaptation. We exclude the videos in the PA-HMDB51 from the HMDB51 to get the training set. Similar to the UCF101 experiments, the target utility task $\mathcal{T}$ (\ie, action recognition) can be taken as a video classification problem with 51 classes, and the privacy budget task $\mathcal{B}$ (\ie, privacy attribute prediction) can be taken as a multi-label image classification task with two classes for each privacy attribute label. Notably, although PA-HMDB51 has provided concrete multi-class labels with specific privacy attribute classes, we convert them into binary labels during testing. 
For example, for ``gender'' attribute, we have provided ground truth labels ``male,'' ``female,'' ``coexisting,'' and ``cannot tell,'' 
but we only use ``can tell'' and ``cannot tell'' in our experiments, via combining ``male,'' ``female'' and ``coexisting'' into the one class of ``can tell.''
This is because we must keep the testing protocol on PA-HMDB51 consistent with the training protocol on VISPR (a multi-label, ``either-or'' type \emph{binary} classification task, so that our models cross-trained on UCF101-VISPR can be evaluated directly.
We hope to extend training to PA-HMDB51 in the future so that the privacy budget task can be formulated and evaluated as a multi-label classification problem.

% The inputs $X$ to our framework are clips with shape $[T,W,H,C]=[16,112,112,3]$, just the same as in the SBU and UCF101 experiments.
All implementation details are identical with the UCF101 case, except that we adjust $th_T=0.7$ and $th_B=0.95$.

\subsubsection{Results and Analysis}

\begin{figure}[!t]
\centering
\includegraphics[width=\columnwidth]{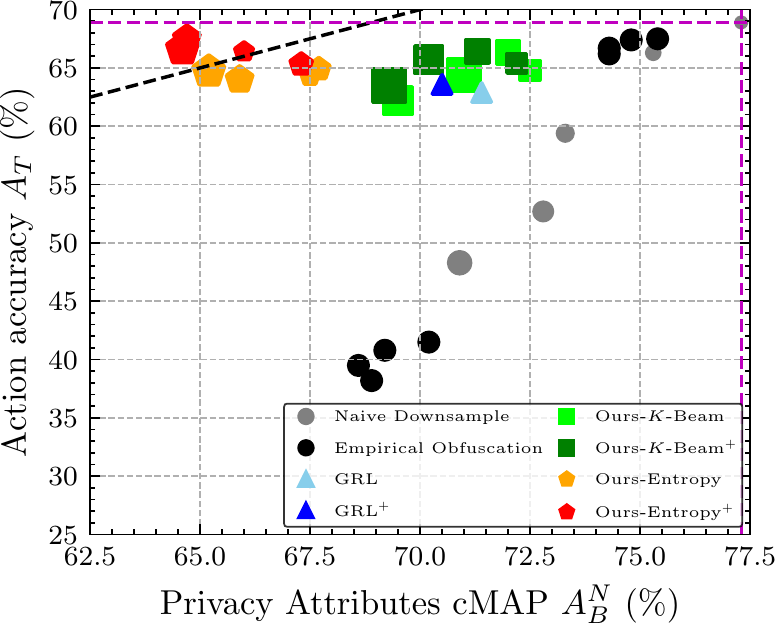}
\vspace{-1.5em}
\caption{The trade-off between privacy budget and action utility on PA-HMDB51 Dataset. For \emph{Naive Downsample} method, a larger marker means a larger downsampling rate is adopted. For \emph{Ours-$K$-Beam} method, a larger marker means a larger $K$ (number of beams) in Algorithm~\ref{alg:kbeam}. For \emph{Ours-Entropy} and \emph{Ours-Entropy (restarting)}, a larger marker means a larger $M$ (number of ensemble models) in Algorithm~\ref{alg:ME2}.
Methods with ``+'' superscript are incorporated with model restarting.
Vertical and horizontal purple dashed lines indicate $A_B^N$ and $A_T$ on the original non-anonymized videos, respectively.
The black dashed line indicates where $A_B^N=A_T$.
Detailed experimental settings and numerical results for each method can be found in Appendix B.
}
% \vspace{-1em}
\label{fig:HMDB51}
\end{figure}

The results on PA-HMDB51 are shown in Figure~\ref{fig:HMDB51}. Our methods achieve much better trade-off between privacy budget and action utility compared with baseline methods.
When $M=4$, our methods can decrease privacy cMAP by around 8\% with little harm to utility accuracy. Overall, the privacy gains are more limited compared to the previous two experiments, because no (re-)training is performed; but the overall comparison trends show the same consistency.

\vspace{+0.25em}
\noindent \textbf{Asymmetrical Privacy Attributes Protection Cost.} 
Different privacy attributes have different protection costs. After applying the learned anonymization optimized by \emph{Ours-Entropy (restarting, $M$=$4$)} on PA-HMDB51, the drop in AP of ``face'' is much more significant than ``gender,'' which indicates that the ``gender'' attribute is much harder to suppress than ``face.'' Such observation agrees that the gender attribute can be revealed by face, body, clothing, and even hairstyle. In future work, we will take such cost asymmetry into account by using a weighted loss combination of different privacy attributes or training dedicated privacy protector for the most informative private attribute.

\vspace{+0.25em}
\noindent \textbf{Human Study on the Privacy Protection of Our Learned Anonymization.}
We use a human study to evaluate the trade-off between privacy budget and action utility achieved by our learned anonymization transform. We take both privacy protection and action recognition into account in the study. We emphasize here that both privacy protection and action recognition are evaluated on video level. There are $515$ videos distributed on $51$ actions in the PA-HMDB51. For each action in the PA-HMDB51, we randomly pick one video for the human study. Among the $51$ selected videos, we only keep $30$ videos to reduce the human evaluation cost. There were $40$ volunteers involved in the human study. In the study, they were asked to label all the privacy attributes and the action type on the raw videos and the anonymized videos. According to the experimental results 
% (shown in Appendix \ref{sec:human_study}), 
(shown in Appendix E), 
the actions in the anonymized videos are still distinguishable to humans, but the privacy attributes are not recognizable at all. This human study further justifies that our learned anonymization transform can protect the privacy and maintain target utility task performance simultaneously.

\section{Conclusion}
We propose an innovative framework to address the newly-established problem of privacy-preserving action recognition. 
To tackle the challenging adversarial learning process, we investigate three different optimization schemes. 
To further tackle the $\forall$ challenge of universal privacy protection, we propose the privacy budget model restarting and ensemble strategies. Both are shown to improve the privacy-utility trade-off further. Various simulations verified the effectiveness of the proposed framework. More importantly, we collect the first dataset for privacy-preserving video action recognition, an effort that we hope could engage a broader community into this research field. 

We note that there is much room to improve the proposed framework before it can be used in practice. For example, the definition of privacy leakage risk is core to the framework. Considering the $\forall$ challenge, current $L_B$ defined with any specific $f_B$ is insufficient; the privacy budget model ensemble could only be viewed as a rough discretized approximation of $\mathcal{P}$. More elaborated ways to approach this $\forall$ challenge may lead to a further breakthrough in achieving the optimization goal.

\section*{Acknowledgement}
Z. Wu, H. Wang, and Z. Wang were partially supported by NSF Award RI-1755701. The authors would like to sincerely thank Scott Hoang, James Ault, and Prateek Shroff for assisting the labeling of the PA-HMDB51 dataset.

% \bibliographystyle{IEEEtran}
% \bibliography{reference} 


\vspace{-3em}
\begin{IEEEbiography}
[{\includegraphics[width=1in,height=1.25in,clip,keepaspectratio]{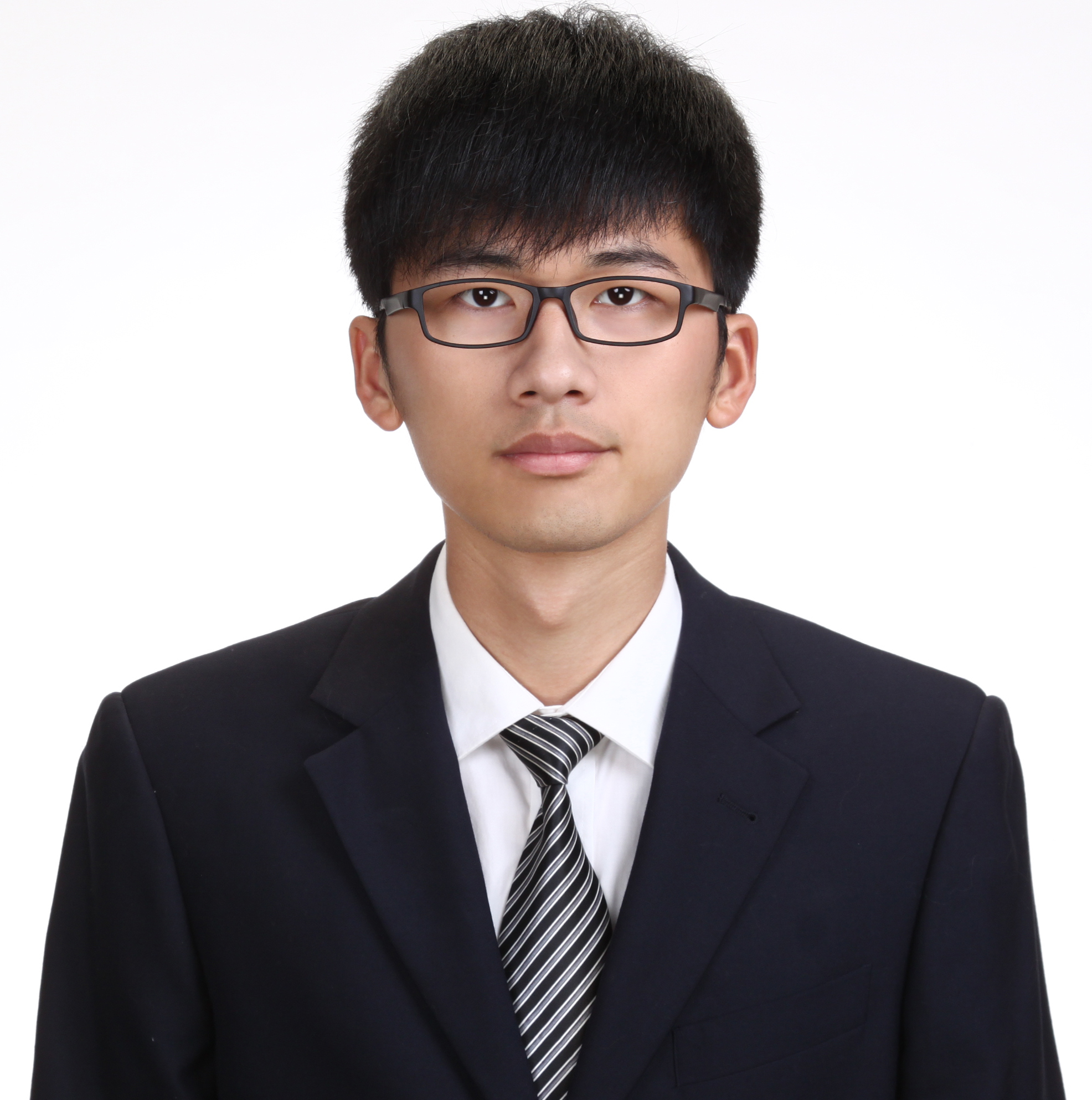}}]
{Zhenyu Wu} received the M.S. and B.E. degrees from the Ohio State University and Shanghai Jiao Tong University, in 2017 and 2015 respectively. He is currently a Ph.D. student at Texas A\&M University, advised by Prof. Zhangyang Wang.
His research interests include privacy/fairness in machine learning, efficient vision, object detection, and adversarial learning. 
\end{IEEEbiography}
\vspace{-3em}

\begin{IEEEbiography}
[{\includegraphics[width=1in,height=1.25in,clip,keepaspectratio]{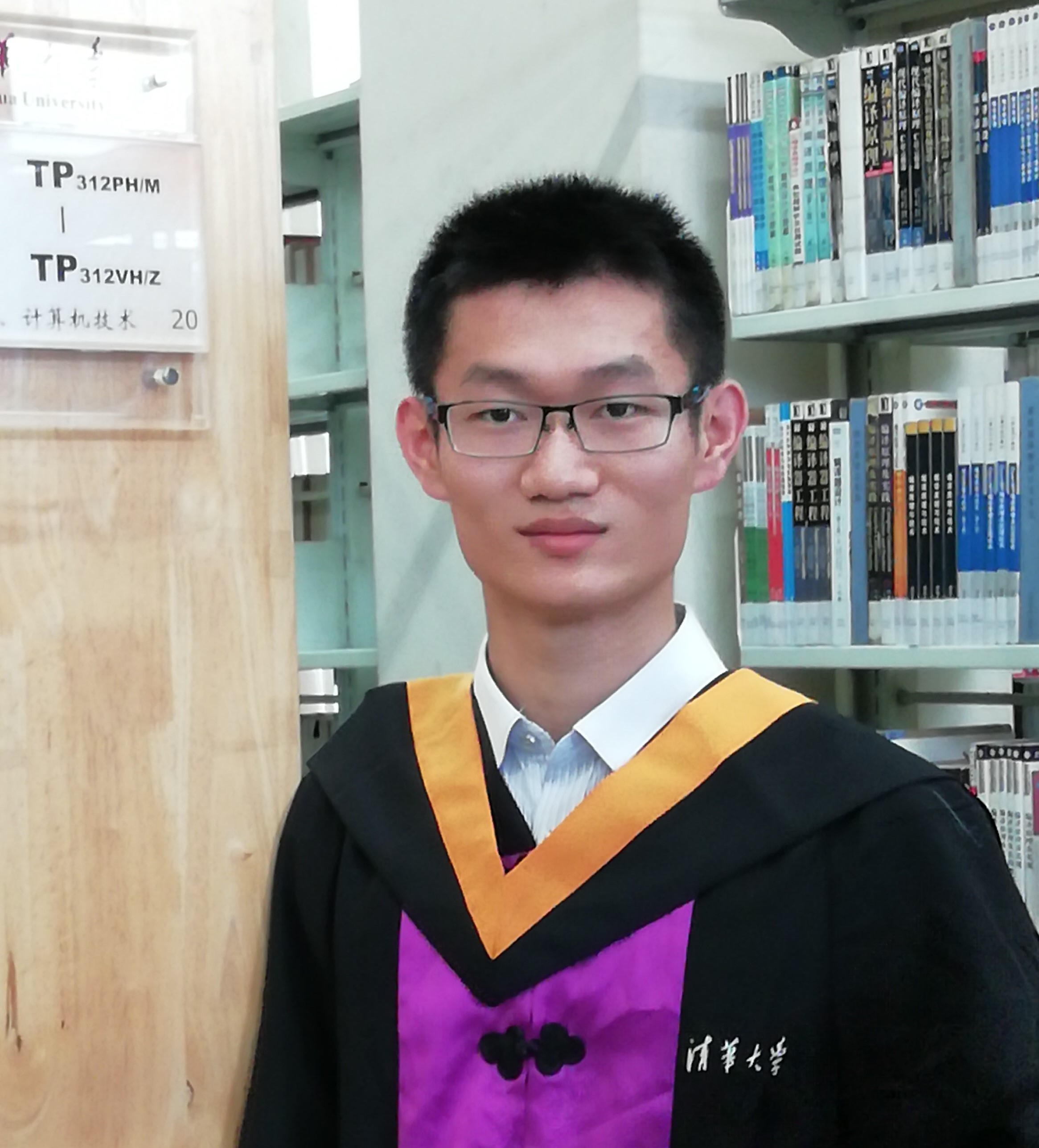}}]
{Haotao Wang} received the B.E. degree in EE from Tsinghua University, China, in 2018. 
He is working toward a Ph.D. degree at the University of Texas at Austin, under the supervision of Prof. Zhangyang Wang. 
His research interests include computer vision and machine learning, especially in fairness/privacy in machine learning, adversarial robustness, and model compression.
\end{IEEEbiography}
\vspace{-3em}

\begin{IEEEbiography}
[{\includegraphics[width=1in,height=1.25in,clip,keepaspectratio]{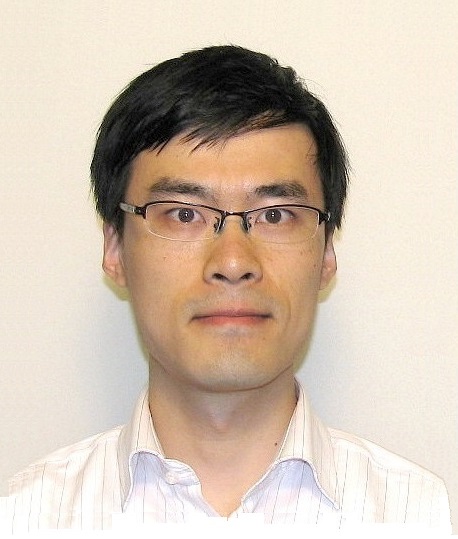}}]
{Zhaowen Wang} received the B.E. and M.S. degrees from Shanghai Jiao Tong  University, China, in 2006 and 2009 respectively, and the Ph.D. degree in ECE from UIUC in 2014. He is currently a Senior Research Scientist with the Creative Intelligence Lab, Adobe Inc. His research focuses on understanding and enhancing images, videos and graphics via machine learning algorithms, with a particular interest in sparse coding and deep learning.
\end{IEEEbiography}
\vspace{-3em}

\begin{IEEEbiography}
[{\includegraphics[width=1in,height=1.25in,clip,keepaspectratio]{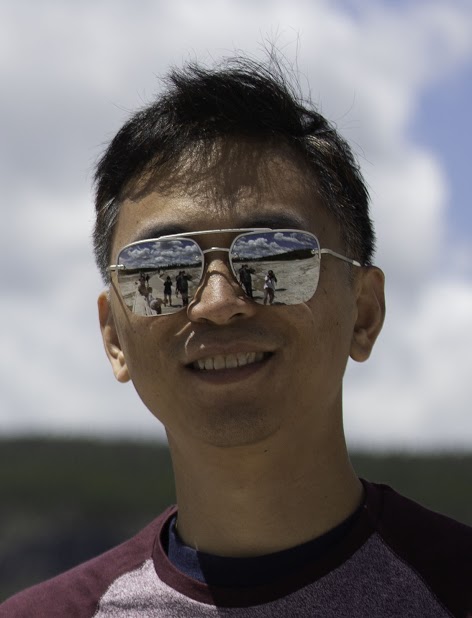}}]
{Hailin Jin} is a Senior Principal Scientist at Adobe Research. He received his M.S. and Ph.D. in EE from WUSTL in 2000 and 2003. Between fall 2003 and fall 2004, he was a postdoc researcher at the CS Department, UCLA. His current research interests include deep learning, computer vision, and natural language processing. His work can be found in many Adobe products, including Photoshop, After Effects, Premiere Pro, and Photoshop Lightroom.
\end{IEEEbiography}
\vspace{-3em}

\begin{IEEEbiography}
[{\includegraphics[width=1in,height=1.25in,clip,keepaspectratio]{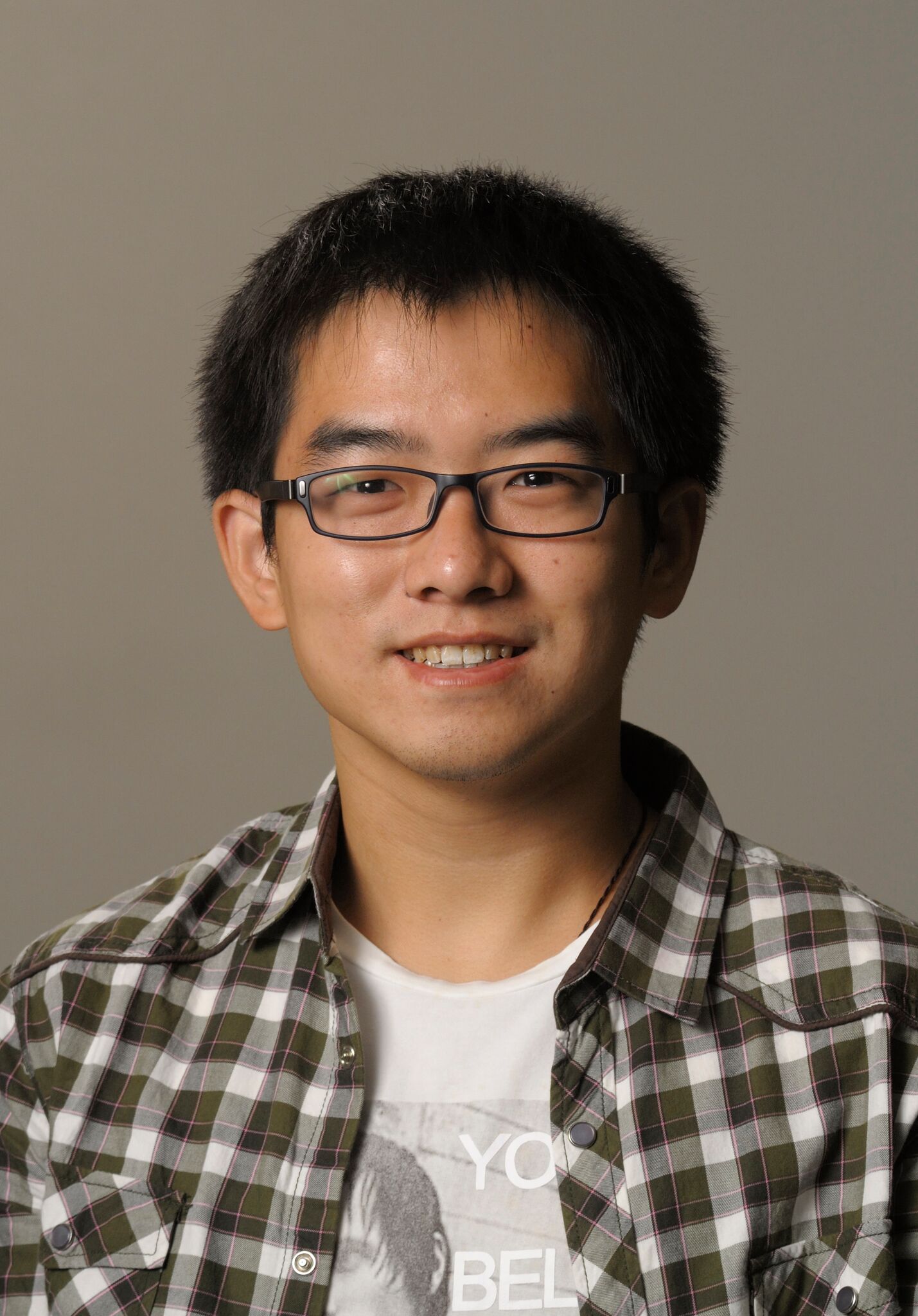}}]
{Zhangyang Wang} is currently an Assistant Professor of ECE at UT Austin. He was an Assistant Professor of CSE, at TAMU, from 2017 to 2020. He received his Ph.D. in ECE from UIUC in 2016, and his B.E. in EEIS from USTC in 2012. Prof. Wang is broadly interested in the fields of machine learning, computer vision, optimization, and their interdisciplinary applications. His latest interests focus on automated machine learning (AutoML), learning-based optimization, machine learning robustness, and efficient deep learning. 
\end{IEEEbiography}

\clearpage
\appendices
% \documentclass[10pt,journal,compsoc]{IEEEtran}

% \usepackage{cite}
% \usepackage[pdftex]{graphicx}
% \usepackage{amsmath}
% \usepackage{amssymb}
% \usepackage{mathtools}
% \usepackage{algorithmic}
% \usepackage[ruled, linesnumbered]{algorithm2e}
% \usepackage{array}
% \usepackage[caption=false,font=footnotesize]{subfig}
% \usepackage{fixltx2e}
% \usepackage{stfloats}
% \usepackage{url}
% \usepackage{multirow}
% \usepackage{xcolor}
% \usepackage{enumitem}
% \usepackage{hyperref}
% \usepackage[center]{caption}

% \DeclareMathOperator*{\argmin}{argmin} % thin space, limits
% \DeclareMathOperator*{\argmax}{argmax} 

% % \bibliographystyle{ieeetr}

% \hyphenation{op-tical net-works semi-conduc-tor}

% \begin{document}
% \title{Supplementary Materials for ``Privacy-Preserving Deep Visual Recognition: \\ An Adversarial Learning Framework \\ and A New Dataset''}
% \author{Zhenyu~Wu*,
%         Haotao~Wang*, 
%         Zhangyang~Wang,
%         Zhaowen~Wang,
%         and~Hailin~Jin% <-this % stops a space
% \IEEEcompsocitemizethanks{\IEEEcompsocthanksitem Zhenyu Wu, Haotao Wang,  and Zhangyang Wang are with the Department
% of Computer Science and Engineering, Texas A\&M University, College Station,
% TX, 77840.\protect\\
% E-mail: \{wuzhenyu\_sjtu,htwang,atlaswang\}@tamu.edu \protect\\
% The first two authors contributed equally to this work. \protect\\
% \IEEEcompsocthanksitem Zhaowen Wang and Hailin Jin are with Adobe Research, San Jose, CA, 95110. E-mail: \{zhawang,hljin\}@adobe.com
% }% <-this % stops an unwanted space
% }

% % make the title area
% \maketitle

\section{GRL and Ours-$K$-Beam with Model Restarting} \label{sec:app-algo}

In this section, we provide the formal descriptions of GRL and Ours-$K$-Beam algorithms with model restarting in Algorithm~\ref{alg:GRL2} and Algorithm~\ref{alg:kbeam2} respectively.

\begin{algorithm}
\SetAlgoLined
Initialize $\theta_A$, $\theta_{T}$ and $\theta_{B}$\;
\For{$t \gets 1$ \KwTo max\_iter}
{
    \If{$t\equiv 0 \pmod{rstrt\_iter}$}
    {
        Reinitialize $\{\theta_B\}$ \\
    }
    Update $\theta_A$ using Eq.~(\ref{eq:GRL-D}) \\
    \While{$Acc(\mathcal{X}^v,\mathcal{Y}_T^v)$ $\leq th_T$ }
    {
        Update $\theta_T$ using Eq.~(\ref{eq:GRL-T})
    }
    \While{$Acc(\mathcal{X}^t,\mathcal{Y}_B^t)$ $\leq th_B$}
    {
        Update $\theta_{B}$ using Eq.~(\ref{eq:GRL-B})
    }
}
\caption{GRL algorithm (with model restarting)}
 \label{alg:GRL2}
\end{algorithm}

\begin{algorithm}
\SetAlgoLined
Initialize $\theta_A$, $\theta_{T}$ and $\{\theta_B^i\}_{i=1}^K$\;
\For{$t \gets 1$ \KwTo max\_iter}
{
    \If{$t\equiv 0 \pmod{rstrt\_iter}$}
    {
        Reinitialize $\{\theta_{B}^{i}\}_{i=1}^{K}$ \\
    }
    \textbf{/*$L_T$ step:*/} \\
    \While{$Acc(\mathcal{X}^v,\mathcal{Y}_T^v)$ $\leq th_T$ }
    {
        Update $\theta_T$, $\theta_A$ using Eq.~(\ref{eq:kbeam-LT})
    }
    \textbf{/*$L_B$ \emph{Max} step:*/} \\
    Update $j$ using Eq.~(\ref{eq:kbeam-LB-argMAX}) \\
    \For{$t_2\gets1$ \KwTo d\_iter}
    {
        Update $\theta_A$ using Eq.~(\ref{eq:kbeam-LB-MAX})
    }
    \textbf{/*$L_B$ \emph{Min} step:*/} \\
    \For{$i\gets1$ \KwTo K}
    {
        \While{$Acc(\mathcal{X}^t,\mathcal{Y}_B^t)$ $\leq th_B$}
        {
            Update $\theta_B^i$ using Eq.~(\ref{eq:kbeam-LB-MIN})
        }
    }
}
\caption{Ours-$K$-Beam algorithm (with model restarting)}
\label{alg:kbeam2}
\end{algorithm}

\section{Detailed numerical results} \label{sec:app-numer-results}
In Table~\ref{tab:num-results}, we provide detailed numerical results reported in Figures~\ref{fig:SBU1},~\ref{fig:UCF101}~and~\ref{fig:HMDB51}.

\begin{table}[h] 
\caption{Detailed numerical results of all the experiments. $A_T$ stands for target utility task (action recognition) accuracy while $A_B^N$ stands for privacy budget prediction performance (accuracy in classification task and cMAP in multi-label classification task). $r$ is the sampling rate for the downsampling baselines. \{box: \textbf{X}, segmentation: \textbf{S}\} $\times$ \{blurring: \textbf{B}, blackening: \textbf{K}\} $\times$ \{face: \textbf{F}, human body: \textbf{D}\} are different empirical obfuscation baselines. $K$ is the number of different sets of budget model parameters tracked by Ours-$K$-Beam. $M$ is the number of ensemble budget models used by Ours-Entropy. Methods with ``$+$'' superscript are incorporated with model restarting.}
\centering
\resizebox{\columnwidth}{!}{%

\begin{tabular}[t]{c|c|c|c|c|c|c|c}
\hline
{} & {} & \multicolumn{2}{c|}{SBU} & \multicolumn{2}{c|}{UCF101} & \multicolumn{2}{c}{HMDB51} \\
\hline
{} & Method & \makecell{$A_T$} & \makecell{$A_B^N$} & \makecell{$A_T$} & \makecell{$A_B^N$} & \makecell{$A_T$} & \makecell{$A_B^N$}\\
\hline
\multirow{5}{*}{Downsample} & \makecell{$r=1$} & {88.8} & {99.5} & {69.3} & {45.5} & {68.9} & {77.3} \\
& \makecell{$r=2$} & {87.9} & {99.2} & {67.9} & {42.1} & {66.3} & {75.3} \\
& \makecell{$r=4$} & {81.9} & {99.1} & {50.2} & {38.6} &  {59.4} & {73.3} \\
& \makecell{$r=8$} & {74.9} & {97.6} & {43.4} & {34.4} & {52.7} & {72.8} \\
& \makecell{$r=16$} & {64.4} & {97.2} & {32.8} & {31.2} & {48.3} & {70.9} \\
\hline
\multirow{8}{*}{Obfuscation} & XKF & 88.5 & 98.9 & 64.2 & 41.3 & 66.7 & 74.3 \\
& XKD & 24.6 & 21.4 & 45.7 & 24.7 & 38.2 & 68.9 \\
& SKF & 88.4 & 98.4 & 64.5 & 41.0 & 67.4 & 74.8 \\
& SKD & 44.6 & 49.5 & 46.5 & 30.7 & 41.5 & 70.2 \\
& XBF & 88.3 & 98.5 & 63.2 & 40.7 & 66.2 & 74.3 \\
& XBD & 58.3 & 43.4 & 46.5 & 29.2 & 40.8 & 69.2 \\
& SBF & 88.6 & 98.5 & 64.1 & 40.5 & 67.5 & 75.4 \\
& SBD & 54.2 & 36.3 & 47.6 & 31.5 & 39.5 & 68.6 \\
\hline
\multirow{2}{*}{Ours-GRL} & \makecell{$\text{GRL}$} & {80.9} & {85.4} & {63.6} & {35.9} & {62.9} & {71.4} \\
& \makecell{$\text{GRL}^+$} & {82.8} & {84.3} & {65.4} & {36.3} & {63.3} & {70.5} \\
\hline
\multirow{6}{*}{Ours-$K$-Beam} & \makecell{$K=1$} & {75.5} & {85.9} & {64.3} & {41.7} & {64.8} & {72.5} \\
& \makecell{$K=1^+$} & {78.4} & {83.9} & {65.7} & {40.6} & {65.4} & {73.2} \\
& \makecell{$K=2$} & {80.5} & {87.7} & {64.5} & {40.4} & {64.4} & {71.0} \\
& \makecell{$K=2^+$} & {84.5} & {80.6} & {65.5} & {39.5} & {63.5} & {69.3} \\
& \makecell{$K=4$} & {80.9} & {88.4} & {65.9} & {39.7} & {62.2} & {69.5} \\
& \makecell{$K=4^+$} & {82.6} & {81.5} & {65.9} & {38.6} & {65.7} & {70.2} \\
& \makecell{$K=8$} & {68.2} & {80.2} & {66.2} & {38.4} & {66.3} & {72.0} \\
& \makecell{$K=8^+$} & {74.7} & {80.3} & {67.0} & {37.5} & {66.4} & {71.3} \\
\hline
\multirow{8}{*}{Ours-Entropy} & \makecell{$M=1$} & {83.2} & {73.6} & {64.3} & {40.7} & {65.2} & {70.1} \\
& \makecell{$M=1^+$} & {81.7} & {60.5} & {66.6} & {38.4} & {66.0} & {70.4} \\
& \makecell{$M=2$} & {84.1} & {72.4} & {64.9} & {39.9} & {65.9} & {70.4} \\
& \makecell{$M=2^+$} & {82.6} & {57.9} & {67.5} & {36.3} & {67.3} & {70.3} \\
& \makecell{$M=4$} & {82.7} & {73.0} & {64.0} & {38.6} & {67.7} & {71.5} \\
& \makecell{$M=4^+$} & {78.0} & {54.6} & {65.3} & {35.2} & {64.7} & {71.6} \\
& \makecell{$M=8$} & {80.8} & {67.2} & {64.7} & {38.3} & {67.5} & {70.5} \\
& \makecell{$M=8^+$} & {82.2} & {47.7} & {66.4} & {34.5} & {64.6 } & {70.1} \\
\hline
\end{tabular}
}
\label{tab:num-results}
\end{table}

% \section{Degraded Frames Visualization}
% We provide the visualization of anonymized frames (as discussed in Section 4.3 of the main paper) on UCF101 in Figure~\ref{fig:ucf-visualize}.

\section{Transferability study of privacy attributes between UCF101/HMDB51 and VISPR}\label{sec:transf-priv}

\begin{figure*}[!t]
    \captionsetup[subfigure]{labelformat=empty,justification=centering,farskip=2pt,captionskip=1pt}
    \centering
	\begin{tabular}{cc}
		\subfloat[ApplyLipStick]{\includegraphics[width=0.45\linewidth, height=0.3\linewidth]{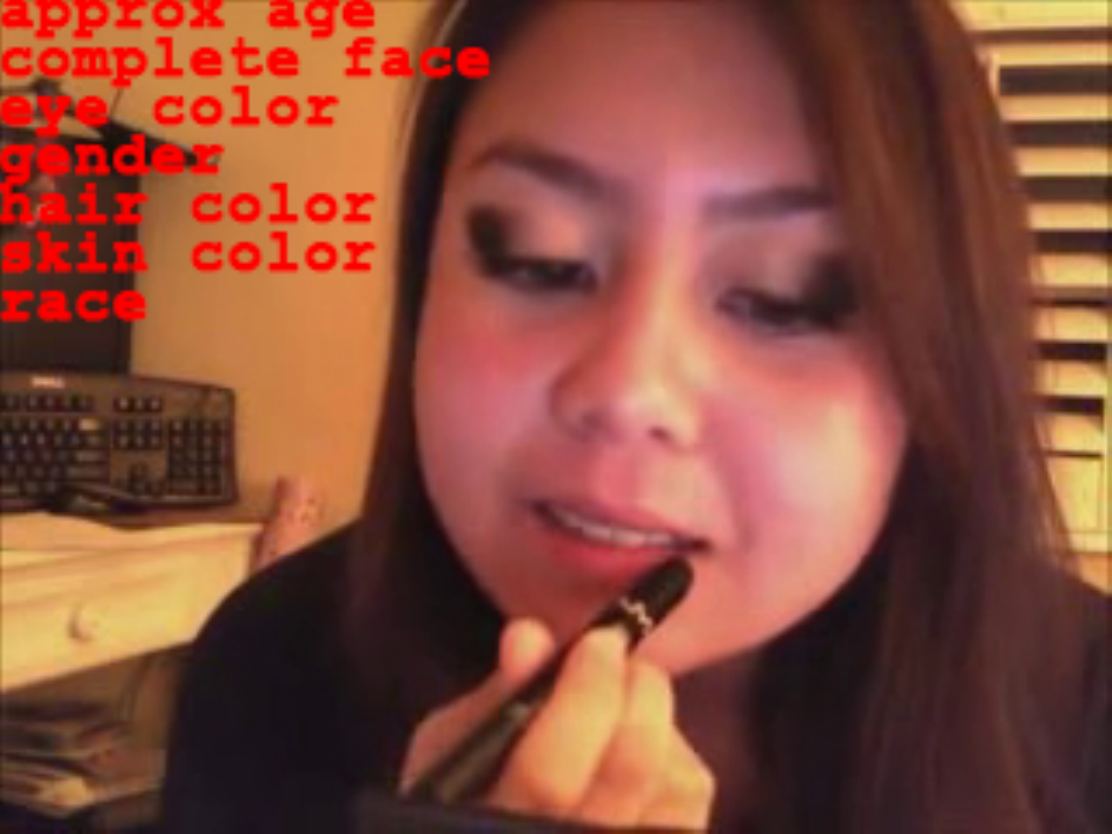}} &
		\subfloat[HandStand]{\includegraphics[width=0.45\linewidth, height=0.3\linewidth]{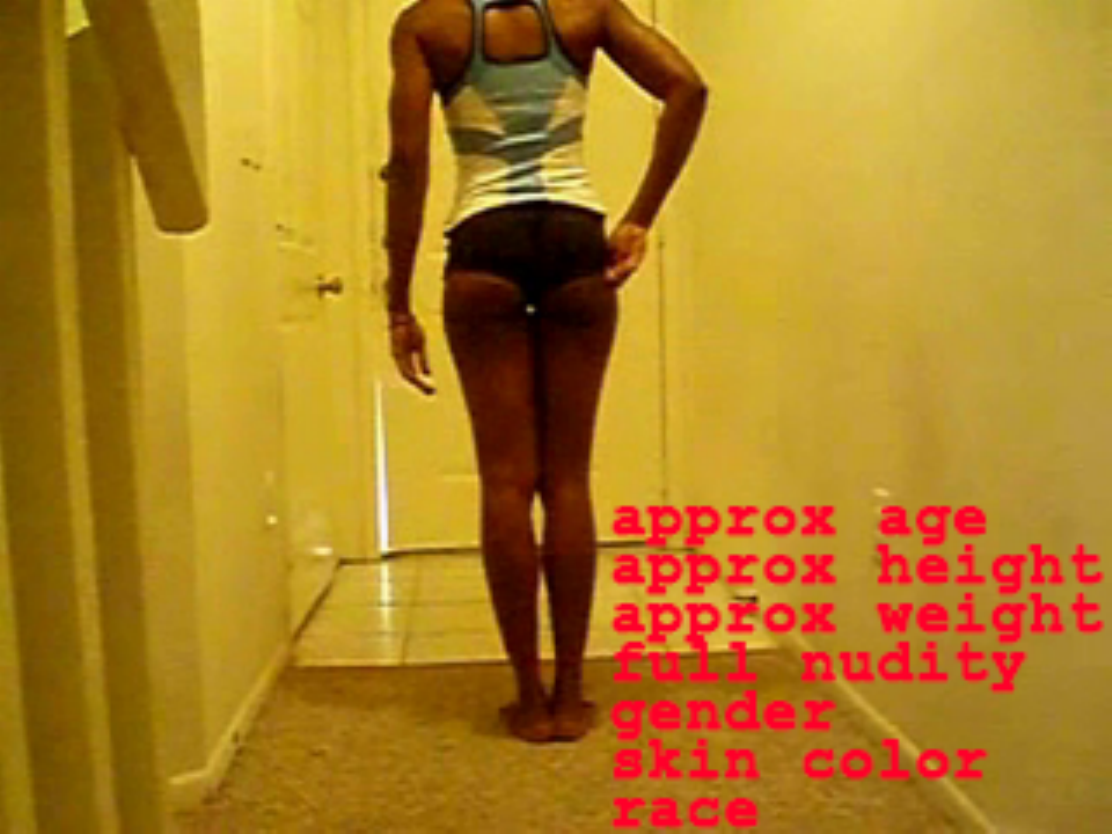}} \\
		\subfloat[Kiss]{\includegraphics[width=0.45\linewidth, height=0.3\linewidth]{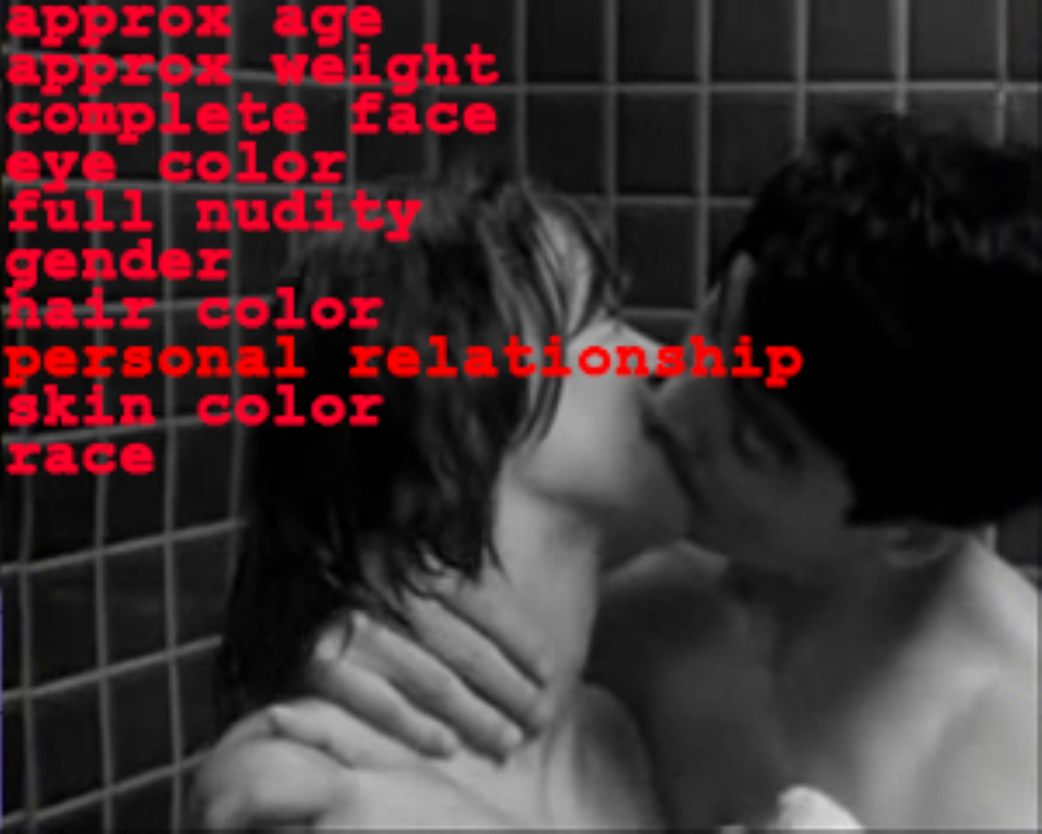}} &
		\subfloat[Pullup]{\includegraphics[width=0.45\linewidth, height=0.3\linewidth]{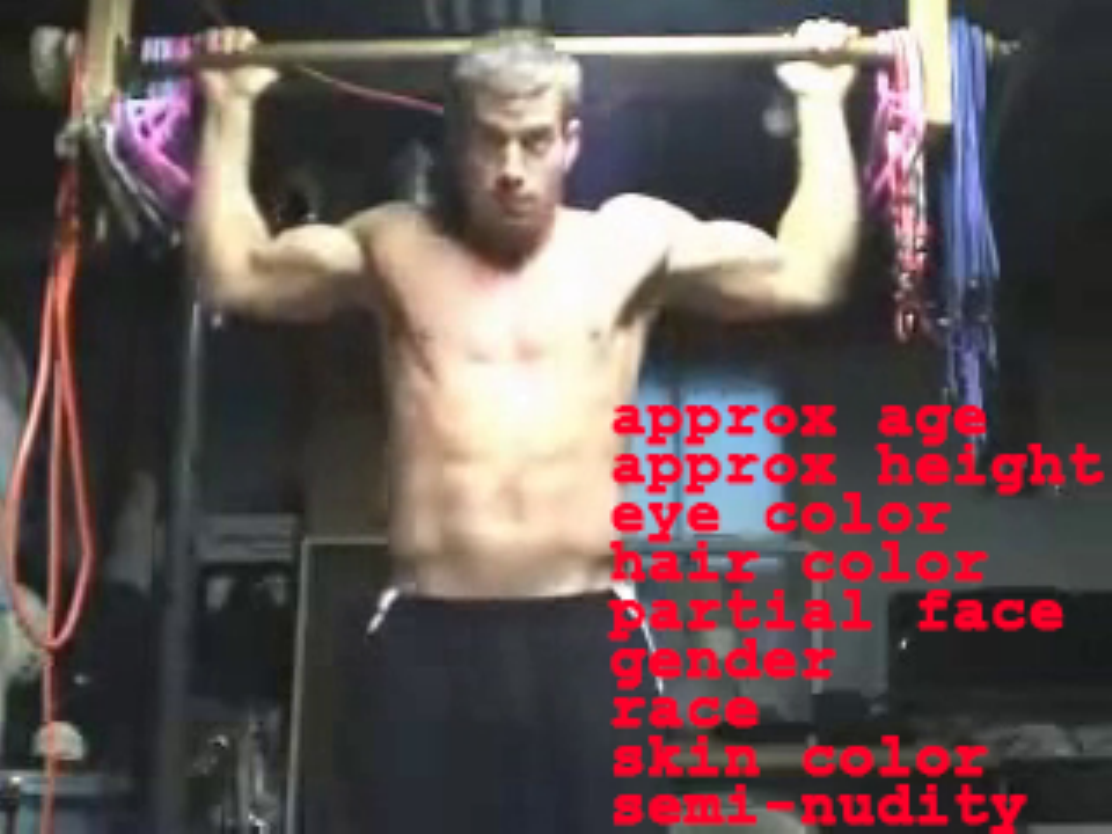}} \\
		\subfloat[Pushup]{\includegraphics[width = 0.45\linewidth, height=0.3\linewidth]{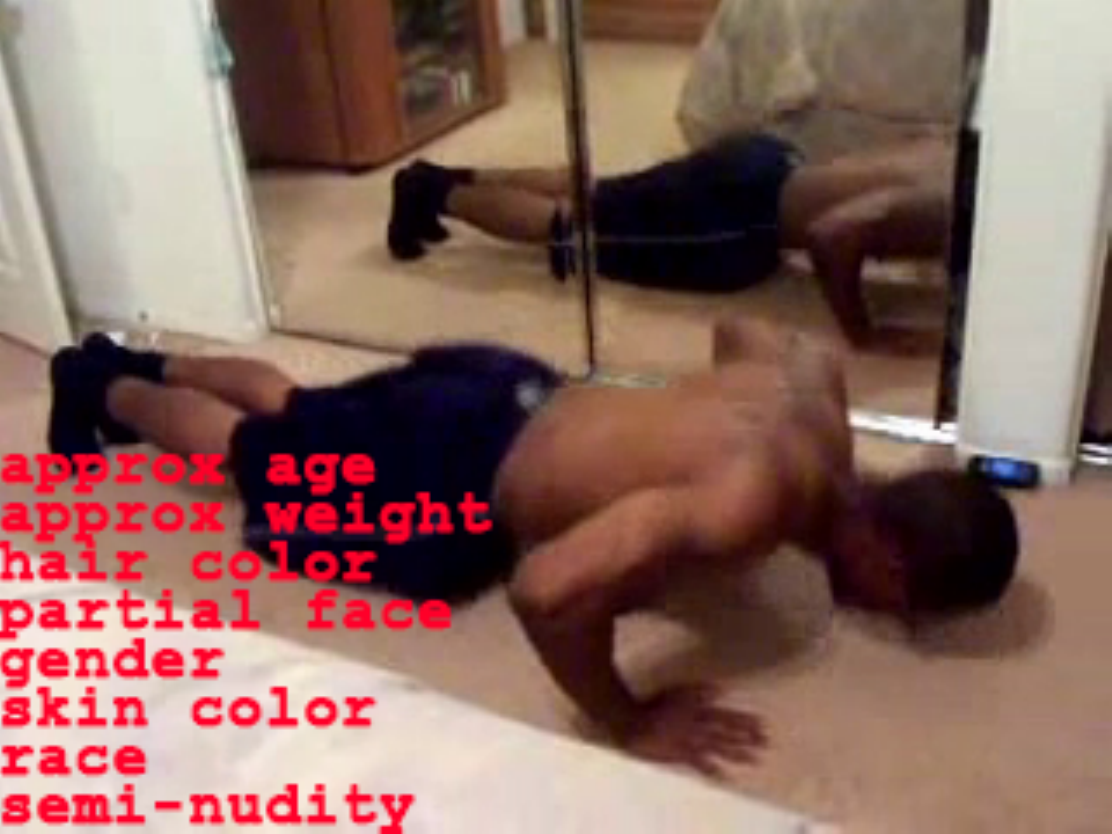}} &
		\subfloat[ShavingBeard]{\includegraphics[width = 0.45\linewidth, height=0.3\linewidth]{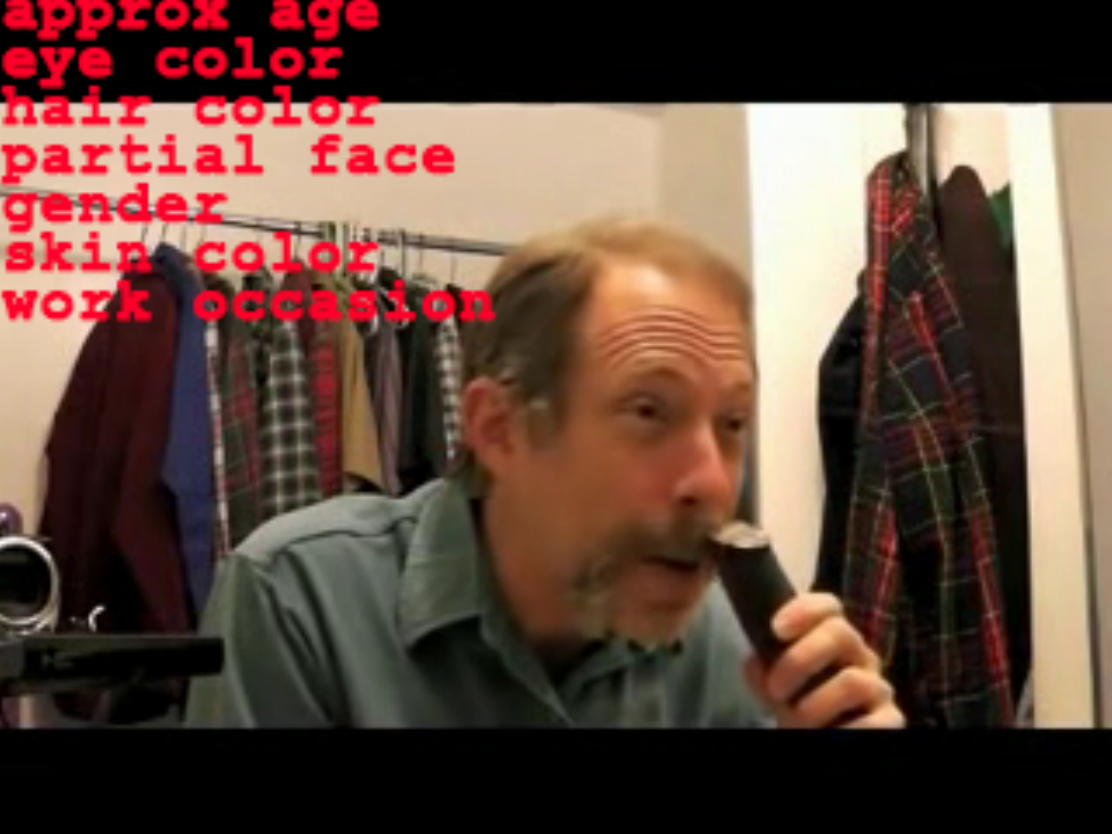}} \\
		\subfloat[Sit]{\includegraphics[width = 0.45\linewidth, height=0.3\linewidth]{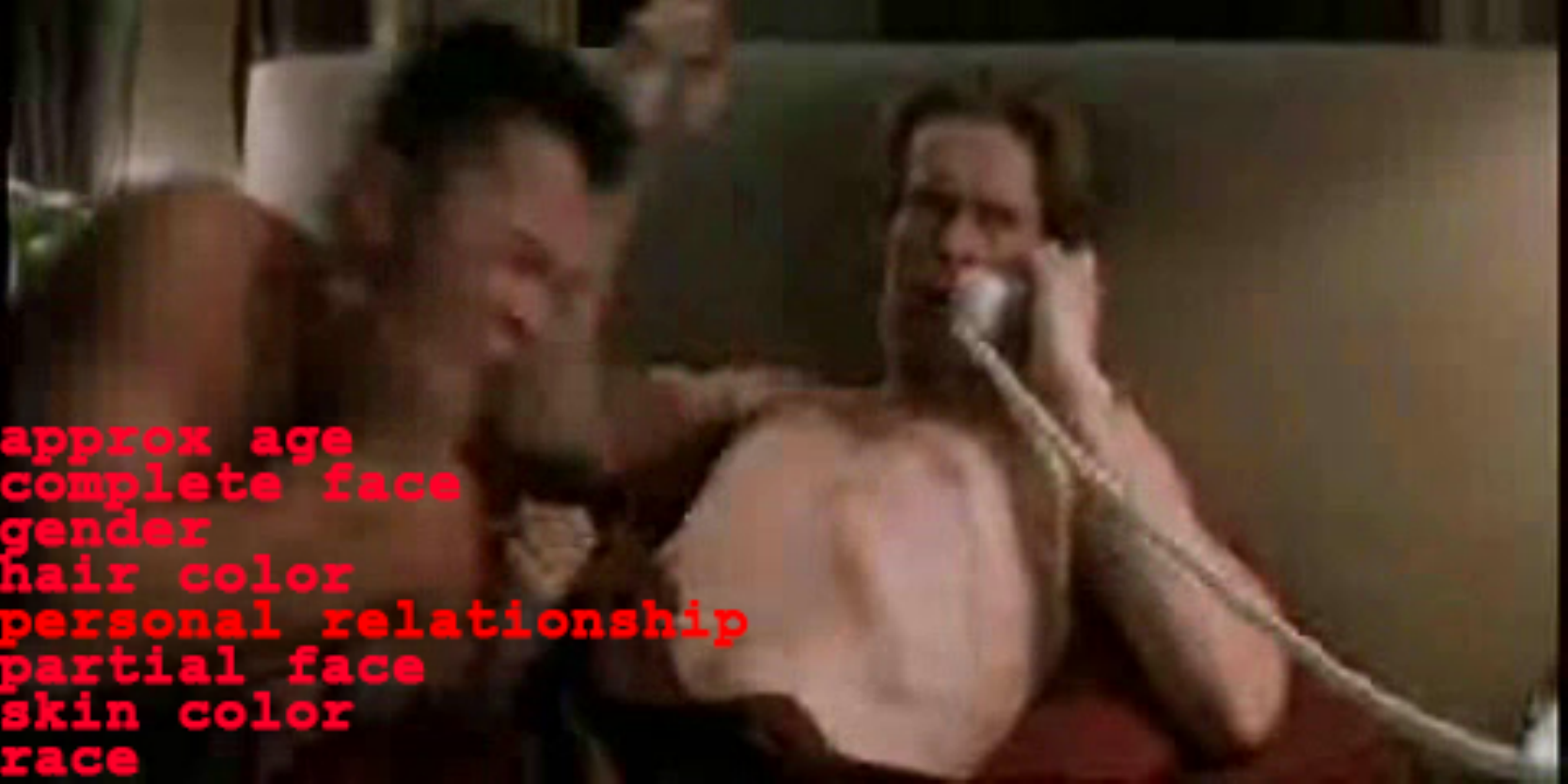}} &
		\subfloat[Sit-up]{\includegraphics[width = 0.45\linewidth, height=0.3\linewidth]{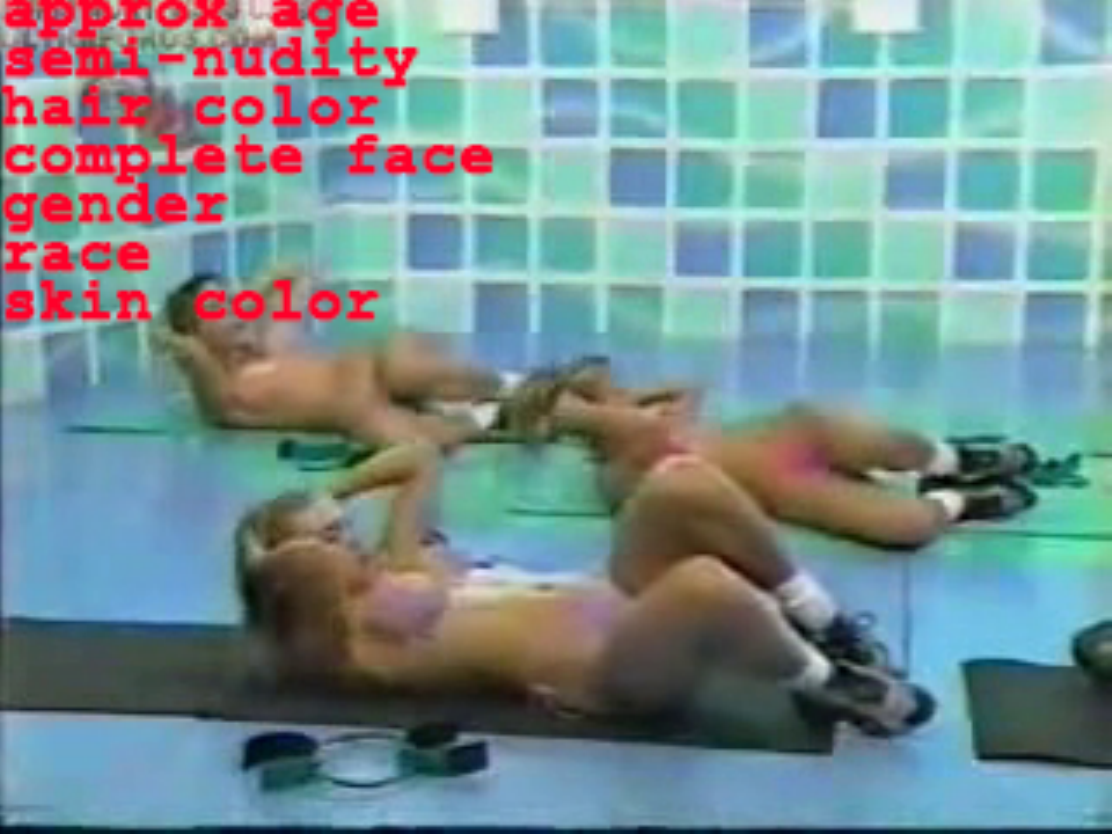}} \\
	\end{tabular}
	\caption{Privacy attributes prediction on selected frames from UCF101 and HMDB51. In each example, the overlayed red text lines denote the privacy attributes (as defined in the VISPR dataset~\cite{orekondy2017towards}) predicted by the privacy prediction model pretrained on VISPR, showing a high risk of privacy leak in videos recording daily activities. The common privacy attributes in daily activities include ``approximate age,'' ``approximate weight,'' ``hair color,'' ``skin color,'' ``partial face,'' ``complete face,'' ``race,'' ``semi-nudity,'' ``gender,'' ``personal relationship,'' and so on.}
	\label{fig:ucf_examples}
\end{figure*}

In Figure~\ref{fig:ucf_examples}, we show some example frames from UCF101 and HMDB51 with privacy attributes predicted using the privacy prediction model pretrained on VISPR. In each example, the overlayed red text lines denote the predicted privacy attributes (as defined in the VISPR dataset~\cite{orekondy2017towards}), showing a high risk of privacy leak in common daily videos. The predicted privacy attributes include ``approximate age,'' ``approximate height,'' ``approximate weight,'' ``semi-nudity,'' ``full-nudity,'' ``partial face,'' ``complete face,'' ``eye color,'' ``hair color,'' ``skin color,'' ``race,'' ``gender,'' ``personal relationship,'' and ``work occasion.'' We qualitatively examine a large number of UCF101 frames and determine that privacy attributes prediction is highly reliable. Such high reliability in prediction justifies our hypothesis that the privacy attributes have good ``transferability'' across UCF101/HMDB51 and VISPR.

\section{More Statistics on PA-HMDB51 Dataset} \label{sec:app-stat}
% \noindent \textbf{Privacy Attribute Distribution}
% Figure \ref{fig:pa-distribution} shows the frame-level label distribution of all five privacy attributes, which is discussed in Section~\ref{sec:pa-distribution} of main paper.

\begin{figure}[!htb]
\centering
\includegraphics[width=\columnwidth]{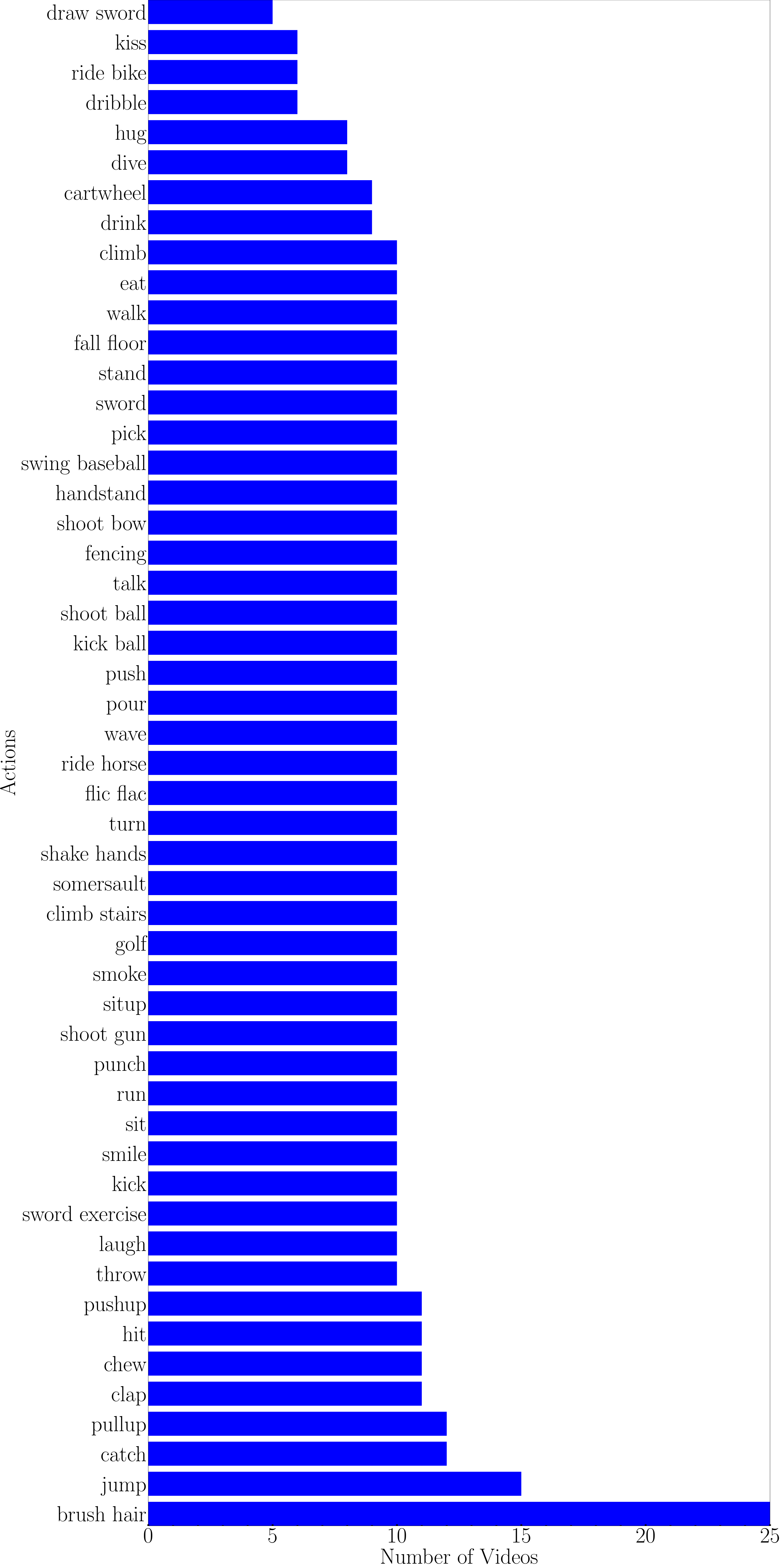}
\caption{Action distribution of PA-HMDB51. Each column shows the number of videos with a certain action. For example, the first column shows there are 25 ``brush hair'' videos in PA-HMDB51 dataset.}
\label{fig:act-dist}
\end{figure}

\noindent \textbf{Action Distribution}
The distribution of action labels (as discussed in Section~\ref{sec:action-pa} of main paper) in PA-HMDB51 are depicted in Figure~\ref{fig:act-dist}, showing a relative class balance.

\begin{figure*}[!t]
\centering
\includegraphics[width=\linewidth]{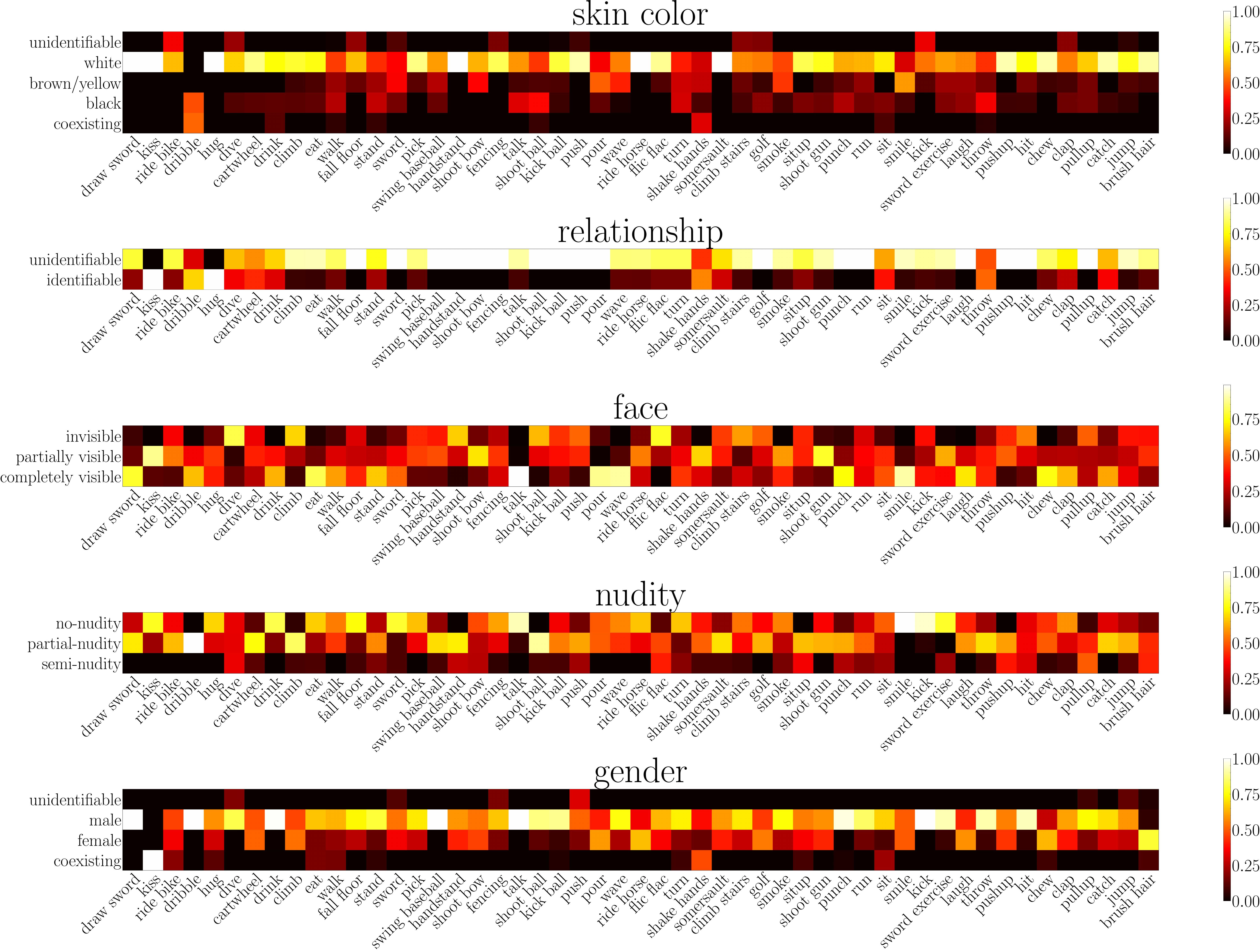}
\caption{Action-attribute correlation in PA-HMDB51 dataset. The color represents ratio of the number of frames of some action containing a specific privacy attribute value \emph{w.r.t.} the total number of frames of the specific action. For example, in the ``relationship'' subplot, the intersection block of row ``identifiable'' and the column ``kiss'' shows the percentage of frames with ``identifiable relationship'' label in all kiss frames.}
\label{fig:act-pa-corl}
\end{figure*}

\noindent \textbf{Action-Attribute Correlation}
We show the correlation between privacy attributes and actions (as discussed in Section~\ref{sec:action-pa} of main paper) in Figure~\ref{fig:act-pa-corl}.

\begin{figure*}[!t]
    \small
    \renewcommand{\arraystretch}{0.5}
	\begin{tabular}{c@{}c@{}c@{}c@{}c@{}c@{}c@{}c@{}c@{}c}
        \includegraphics[width=0.1\linewidth]{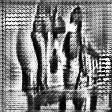} &
        \includegraphics[width=0.1\linewidth]{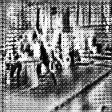} &
        \includegraphics[width=0.1\linewidth]{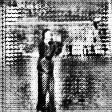} &
        \includegraphics[width=0.1\linewidth]{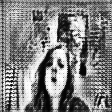} &
        \includegraphics[width=0.1\linewidth]{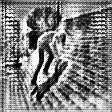} &
        \includegraphics[width=0.1\linewidth]{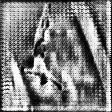} &
        \includegraphics[width=0.1\linewidth]{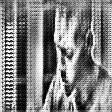} &
        \includegraphics[width=0.1\linewidth]{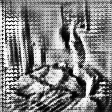} &
        \includegraphics[width=0.1\linewidth]{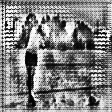} &
        \includegraphics[width=0.1\linewidth]{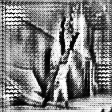} \\
        Brush Hair & Cartwheel & Catch & Chew & Climb & Dive & Eat & Flic-flac & Golf & Handstand
    \end{tabular}
	\begin{tabular}{c@{}c@{}c@{}c@{}c@{}c@{}c@{}c@{}c@{}c}
        \includegraphics[width=0.1\linewidth]{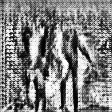} &
        \includegraphics[width=0.1\linewidth]{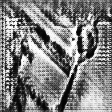} &
        \includegraphics[width=0.1\linewidth]{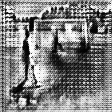} &
        \includegraphics[width=0.1\linewidth]{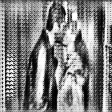} &
        \includegraphics[width=0.1\linewidth]{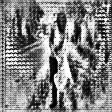} & 
        \includegraphics[width=0.1\linewidth]{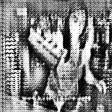} &
        \includegraphics[width=0.1\linewidth]{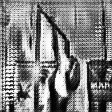} &
        \includegraphics[width=0.1\linewidth]{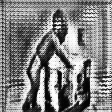} &
        \includegraphics[width=0.1\linewidth]{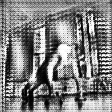} &
        \includegraphics[width=0.1\linewidth]{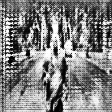} \\
        Hit & Jump & Kick & Kiss & Pick & Pour & Pullup & Push & Pushup & Run
    \end{tabular}
	\begin{tabular}{c@{}c@{}c@{}c@{}c@{}c@{}c@{}c@{}c@{}c}
        \includegraphics[width=0.1\linewidth]{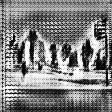} &
        \includegraphics[width=0.1\linewidth]{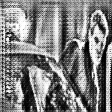} &
        \includegraphics[width=0.1\linewidth]{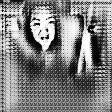} &
        \includegraphics[width=0.1\linewidth]{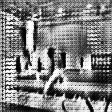} &
        \includegraphics[width=0.1\linewidth]{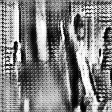} &
        \includegraphics[width=0.1\linewidth]{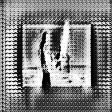} &
        \includegraphics[width=0.1\linewidth]{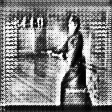} &
        \includegraphics[width=0.1\linewidth]{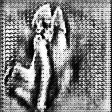} &
        \includegraphics[width=0.1\linewidth]{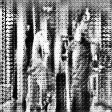} &
        \includegraphics[width=0.1\linewidth]{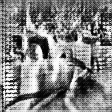} \\
        Shoot ball & Sit & Smile & Somersault & Stand & Sword & Sword Ex. & Throw & Turn & Walk
    \end{tabular}
    \caption{Example frames from PA-HMDB51 used in the human study.}
    \label{fig:human-study-examples}
\end{figure*}

\section{Human Study on Our Learned Anonymization Transform} \label{sec:human_study}
We use a human study to evaluate the privacy-utility trade-off achieved by our learned anonymization transform $f_A^*$. We take both privacy protection and action recognition into account in the study. We emphasize here that both privacy protection and action recognition are evaluated on video level.

\noindent \textbf{Experiment Setting} There are $515$ videos distributed on $51$ actions in PA-HMDB51. For each action in PA-HMDB51, we randomly pick one video for human study. Among the $51$ selected videos, we only keep $30$ videos to reduce the human evaluation cost. The center frame of these $30$ videos are shown in Figure~\ref{fig:human-study-examples}. There are $40$ volunteers involved in the human study. In the study, they were asked to label \textit{all the privacy attributes} and \textit{the action type} on the raw videos and the anonymized videos. The guideline for labeling the privacy attributes in the human study is listed below:
\begin{itemize}[leftmargin=*]
\item gender: the person(s)' gender can be told;
\item nudity: the persons is/are in semi-nudity (wearing shorts/skirts or naked to the waist);
\item relationship: relationships (such as friends, couples, etc.) between/among the actors/actress can be told;
\item face: more than $10\%$ of the face is visible;
\item skin color: the skin color of the person(s) can be told;
\item no privacy attributes found: you cannot tell any privacy attribute.
\end{itemize}
All the experimental results of multi-label privacy attributes prediction in the human study are shown in Table~\ref{raw}. Table~\ref{raw} shows the human study results on the raw videos, the anonymized videos, and random guess. We use the same notations in the paper. $A_T$ stands for action recognition accuracy, and $A_B$ stands for the multi-label privacy attributes prediction macro-F1 score. We use $A_T^r$, $A_T^a$, and $A_T^g$ to represent the action recognition accuracy on the raw videos, on the anonymized videos, and by random guess. Likewise, we use $A_B^r$, $A_B^a$ and $A_B^g$ to represent the macro-F1 score of the privacy attributes prediction on the raw videos, on the anonymized videos, and by random guess. $A_B^a$ ($0.38$) is much lower than $A_B^r$ ($0.97$) and close to $A_B^g$ ($0.51$), justifying the good privacy protection of our learned $f_A^*$. $A_T^a$ ($0.5783$) is comparable to $A_B^r$ ($0.9616$) and significantly higher than $A_B^g$ ($0.0333$), justifying the good target utility preserving of our learned $f_A^*$.

% \begin{table*}[!htb]
% \footnotesize
% \begin{minipage}[t]{0.28\textwidth} \centering

\begin{table}
\centering
\caption{Human study on the raw videos and the anonymized videos. Random guess baseline is also provided.``P,'' ``R,'' ``F1'' stand for precision, recall and F1-Score.} % title of Table
\resizebox{\columnwidth}{!}{%
\begin{tabular}{c|c|c|c|c|c|c|c|c|c} % centered columns (4 columns)
\hline
{} & \multicolumn{3}{c|}{Raw Videos} & \multicolumn{3}{c|}{Anonymized Videos} & \multicolumn{3}{c}{Random Guess} \\
\hline
{} & P & R & F1 & P & R & F1 & P & R & F1\\
\hline
Skin Color & 0.98 & 1.00 & 0.99 & 0.98 & 0.12 & 0.21 & 0.51 & 0.94 & 0.66\\
\hline
Face  & 0.97 & 0.99 & 0.98 & 0.94 & 0.40 & 0.56 & 0.47 & 0.66 & 0.55\\ 
\hline
Gender & 0.98 & 1.00 & 0.99 & 0.94 & 0.36 & 0.52 & 0.49 & 0.91 & 0.64\\ 
\hline
Nudity & 0.99 & 0.99 & 0.99 & 0.61 & 0.09 & 0.16 & 0.48 & 0.51 & 0.49\\ 
\hline
Relationship & 0.97 & 0.88 & 0.92 & 0.47 & 0.39 & 0.43 & 0.49 & 0.14 & 0.22\\
\hline
\hline
Micro-Avg & 0.98 & 0.99 & 0.98 & 0.86 & 0.25 & 0.39 & 0.49 & 0.64 & 0.55\\
\hline
Macro-Avg & 0.98 & 0.97 & 0.97 & 0.79 & 0.27 & 0.38 & 0.49 & 0.63 & 0.51\\
\hline
Weighted-Avg & 0.98 & 0.99 & 0.98 & 0.88 & 0.25 & 0.37 & 0.49 & 0.64 & 0.52\\
\hline
Samples-Avg & 0.98 & 0.99 & 0.98 & 0.45 & 0.24 & 0.30 & 0.49 & 0.61 & 0.51\\
\hline
\end{tabular}
}
\label{raw}
\end{table}

% \begin{figure*}[!t]
% 	\centering
% 	\subfloat[Face]
% 	{
%     	\includegraphics[width=0.42\linewidth]{human-study/confusion-matrix/Face.png}
% 	}
% 	\subfloat[Gender]
% 	{
% 	    \includegraphics[width=0.42\linewidth]{human-study/confusion-matrix/Gender.png}
% 	} \\
% 	\subfloat[Nudity]
% 	{
%     	\includegraphics[width=0.42\linewidth]{human-study/confusion-matrix/Nudity.png}
% 	}
% 	\subfloat[Relationship]
% 	{
% 	    \includegraphics[width=0.42\linewidth]{human-study/confusion-matrix/Relationship.png}
% 	} \\
%     \subfloat[Skin color]
% 	{
%     	\includegraphics[width=0.42\linewidth]{human-study/confusion-matrix/Skin Color.png}
% 	}
% 	\caption{The confusion matrices on the 5 attributes annotated on PA-HMDB51}
% 	\label{fig:confusion-matrices}
% \end{figure*}

% \end{document}

\end{document}